\documentclass[11pt]{article}
\usepackage[margin=1in]{geometry}

\usepackage{tikz-cd}
\usepackage{multicol}

\usepackage{anyfontsize}
\usepackage{colortbl}
\usepackage{xcolor}
\usepackage{steinmetz}
\usepackage{graphicx}
\usepackage{amsmath}
\usepackage{amssymb}
\usepackage{booktabs}
\usepackage{placeins}
\usepackage{float}

\title{Multi-Stage Phase-Aware Self-Supervised Reconstruction for Multi-Echo Gradient-Echo MRI}
\author{\begin{minipage}{0.95\textwidth}\centering
Juan Liu\textsuperscript{1} \quad Nian Wang\textsuperscript{1,2,3}\\[0.5em]
\small \textsuperscript{1}Advanced Imaging Research Center, University of Texas Southwestern Medical Center, USA; \\
\textsuperscript{2}Department of Biomedical Engineering, University of Texas Southwestern Medical Center, USA; \\
\textsuperscript{3}Peter O'Donnell Jr. Brain Institute, University of Texas Southwestern Medical Center, USA. \\
Correspondence: Nian Wang (\texttt{nian.wang@utsouthwestern.edu})
\end{minipage}}
\date{}

\begin{document}
\maketitle

\begin{abstract}
\textbf{Purpose:} To develop a phase-aware self-supervised framework for accelerated high-resolution multi-echo gradient-echo (mGRE) MRI that preserves anatomical detail and complex echo information for quantitative mapping.

\textbf{Methods:} We developed multi-stage phase-aware self-supervised reconstruction (MSPR-GRE), combining progressive k-space expansion with phase estimation and refinement. Sequential unrolled networks reconstruct complex multi-echo images and update echo-dependent phase priors from intermediate reconstructions. Later stages use a relaxation-informed temporal subspace for acquisitions with at least four echoes, whereas shorter echo trains are reconstructed directly in the image domain. Networks are trained sequentially using losses on held-out acquired k-space samples, without fully sampled training targets. Evaluation included two ex vivo mouse-brain datasets acquired at 9.4 T (30-$\mu$m eight-echo and 18-$\mu$m dual-echo) and retrospectively undersampled 7 T AHEAD human-brain data. Supervised AHEAD experiments examined the contributions of phase-aware reconstruction and multi-stage refinement.

\textbf{Results:} MSPR-GRE reduced visible artifacts, particularly in regions exhibiting rapid spatial phase variation, and preserved fine anatomical structures in mouse-brain reconstructions. Across five AHEAD test datasets at nominal acceleration factors of R = 4, 8, and 12, MSPR-GRE generally improved metrics for magnitude images, T2* maps, and QSM, with larger benefits at higher acceleration. At R = 12, fourth-echo magnitude NRMSE was $12.09 \pm 0.60\%$ with MSPR-GRE versus $21.55 \pm 1.12\%$ with SSDU. Ablation results showed qualitative benefits of phase refinement; supervised experiments further supported multi-stage phase-aware reconstruction.

\textbf{Conclusion:} MSPR-GRE improves accelerated mGRE reconstruction and quantitative-map agreement with reference data without requiring fully sampled training targets. Further validation is needed to assess generalizability and phase-estimation accuracy.
\end{abstract}

\section{Introduction}

Multi-echo gradient-echo (mGRE) magnetic resonance imaging (MRI) captures complex signals across echo times to characterize tissue relaxation and magnetic susceptibility. Magnitude decay supports $R_2^*/T_2^*$ mapping, while phase evolution enables susceptibility-weighted imaging (SWI) \cite{haacke2004susceptibility} and quantitative susceptibility mapping (QSM) \cite{wang2015qsm}. These contrasts provide information about tissue microstructure and composition \cite{caan2019mp2rageme}; reliable quantitative analysis therefore requires accurate reconstruction of both magnitude and phase.

High-resolution three-dimensional (3D) mGRE is challenging because small voxels reduce signal-to-noise ratio (SNR), $T_2^*$ decay attenuates later echoes, and long scans increase sensitivity to motion and system drift. Parallel imaging (PI) \cite{pruessmann1999sense,griswold2002generalized} and compressed sensing (CS) \cite{lustig2008compressed} accelerate acquisition through k-space undersampling, but reconstruction methods must preserve fine anatomy and complex signal evolution. Spatially varying phase from static-field ($B_0$) inhomogeneity and tissue susceptibility further complicates reconstruction. At high field, rapid phase evolution can cause phase wrapping and intravoxel dephasing, particularly near air--tissue interfaces. These phase variations carry susceptibility information but make accurate reconstruction from undersampled data more difficult.

Optimization-based methods exploit spatial and temporal correlations in undersampled data. Locally low-rank (LLR) regularization has been applied to multi-echo GRE with off-resonance phase evolution \cite{grosser2020llr}. Model-based approaches further incorporate signal evolution into reconstruction, as demonstrated in fetal-brain $R_2^*$ mapping using multi-echo GRE \cite{wang2025rapid}. Temporal-subspace methods reduce dimensionality by representing echo trains with temporal basis functions and spatial coefficient maps \cite{tamir2017t2shuffling}. However, rapid phase evolution can undermine low-rank approximations unless adequately modeled. Echo-planar time-resolved imaging (EPTI) and extensions incorporate phase evolution and $B_0$ estimation or updating into subspace reconstruction \cite{dong2020epti,dong2021variable,hu2026water}, but remain sensitive to model accuracy, initialization, and noise.

Deep-learning (DL) reconstruction combines learned regularization with physics-based data consistency \cite{hammernik2018variational,aggarwal2019modl}. LARO and mcLARO exploit inter-echo information through feature fusion and learned acquisition strategies for mGRE \cite{zhang2023laro,zhang2024mclaro}. However, DL reconstruction
remains challenging because spatially varying phase evolution can complicate the learning of shared features across echoes. In particular, rapid off-resonance
phase accumulation may be difficult to distinguish from undersampling artifacts, especially at later echoes with reduced SNR. Without adequate phase handling, these
variations may limit the effectiveness of learned regularization and compromise echo-to-echo consistency.

Fully sampled training references are difficult to acquire in high-resolution MRI because long scans are motion-sensitive. Self-supervised methods instead learn from undersampled data; SSDU partitions acquired k-space into disjoint subsets for data consistency and loss evaluation \cite{yaman2020ssdu}. Zero-DeepSub combines a low-rank 3D-QALAS signal subspace with scan-specific, zero-shot regularization \cite{jun2024zerodeepsub}, while Joint MAPLE combines self-supervision and a physical model for joint $T_1/T_2^*$ mapping from multi-echo, multi-flip-angle GRE data \cite{heydari2024jointmaple}. However, reliable phase estimation during self-supervised mGRE reconstruction remains challenging. In particular, a single initial phase prior derived from undersampled, noisy data may be inaccurate and cannot benefit from later image refinement.

We propose multi-stage phase-aware self-supervised reconstruction for multi-echo GRE (MSPR-GRE), which combines progressive k-space expansion, inter-stage phase refinement, and echo-count-adaptive reconstruction. Up to three sequential stages use increasingly broad k-space support: Stage~1 forms an initial reconstruction and phase estimate from central samples; Stage~2 adds samples and updates the phase prior using early-echo phase-evolution fitting when enough echoes are available; and Stage~3 further expands support and refines the prior from the Stage~2 images. This progressive updating lets phase estimates improve alongside image detail. For acquisitions with at least four echoes, later stages use a relaxation-informed temporal subspace; shorter echo trains are reconstructed in the image domain. All echoes are reconstructed jointly. Stages are trained sequentially with self-supervised losses on held-out complex k-space; previously trained stages are frozen, and held-out samples are excluded from initialization, phase estimation, and data consistency.

The main contributions of MSPR-GRE are summarized as follows:

\begin{itemize}

    \item \textbf{Progressive phase-aware reconstruction.}
    We progressively expand k-space support and refine phase priors from intermediate reconstructions to address rapid phase variation.

    \item \textbf{Echo-count-adaptive reconstruction.}
    We adapt the reconstruction representation to echo count, using a relaxation-informed temporal subspace for longer trains and image-domain reconstruction for shorter trains.

    \item \textbf{Stage-wise self-supervised learning.}
    We train each stage sequentially on held-out complex k-space while freezing preceding stages, without fully sampled references.

\end{itemize}

\section{Methods}

\subsection{Signal and Acquisition Model}

The complex-valued multi-echo gradient-echo signal is modeled using a single-component spoiled GRE approximation. At spatial location $\mathbf{r}$ and echo time $t_e$, the signal is expressed as

\begin{equation}
    x(\mathbf{r},t_e)
    =\rho(\mathbf{r})e^{-R_2^*(\mathbf{r})t_e}
    e^{i[\phi_0(\mathbf{r})+2\pi\Delta f(\mathbf{r})t_e]},
    \label{eq:gre_signal}
\end{equation}

where $\rho$ is the extrapolated signal magnitude at zero echo time, $R_2^*=1/T_2^*$ is the effective transverse relaxation rate, $\phi_0$ is the echo-independent phase offset, and $\Delta f$ is the local off-resonance frequency in Hz, reflecting contributions from background-field inhomogeneity and tissue susceptibility.

Let $N_v$, $N_e$, and $N_c$ denote the numbers of spatial voxels, echoes, and receive coils, respectively. The multi-echo images are represented by $\mathbf{X}\in\mathbb{C}^{N_v\times N_e}$, with each column corresponding to an echo. Let $\Omega$ denote the acquired k-space samples across echoes and coils. For any subset $\Gamma\subseteq\Omega$, the multi-coil encoding operator is defined as $\mathbf{A}_{\Gamma}=\mathbf{U}_{\Gamma}\mathbf{F}\mathbf{S}$, where $\mathbf{U}_{\Gamma}$ is the sampling operator, $\mathbf{F}$ denotes the Fourier transform, and $\mathbf{S}$ represents coil sensitivity encoding. The acquisition model is

\begin{equation}
    \mathbf{y}_{\Gamma}
    =\mathbf{A}_{\Gamma}\mathbf{X}
    +\boldsymbol{\varepsilon}_{\Gamma},
    \label{eq:forward_model}
\end{equation}

where $\mathbf{y}_{\Gamma}$ contains the acquired k-space measurements and $\boldsymbol{\varepsilon}_{\Gamma}$ denotes measurement noise. The adjoint operator $\mathbf{A}_{\Gamma}^H$ maps k-space measurements back to the multi-echo image domain.

\begin{figure}[!t]
    \centering
    \includegraphics[width=\textwidth]{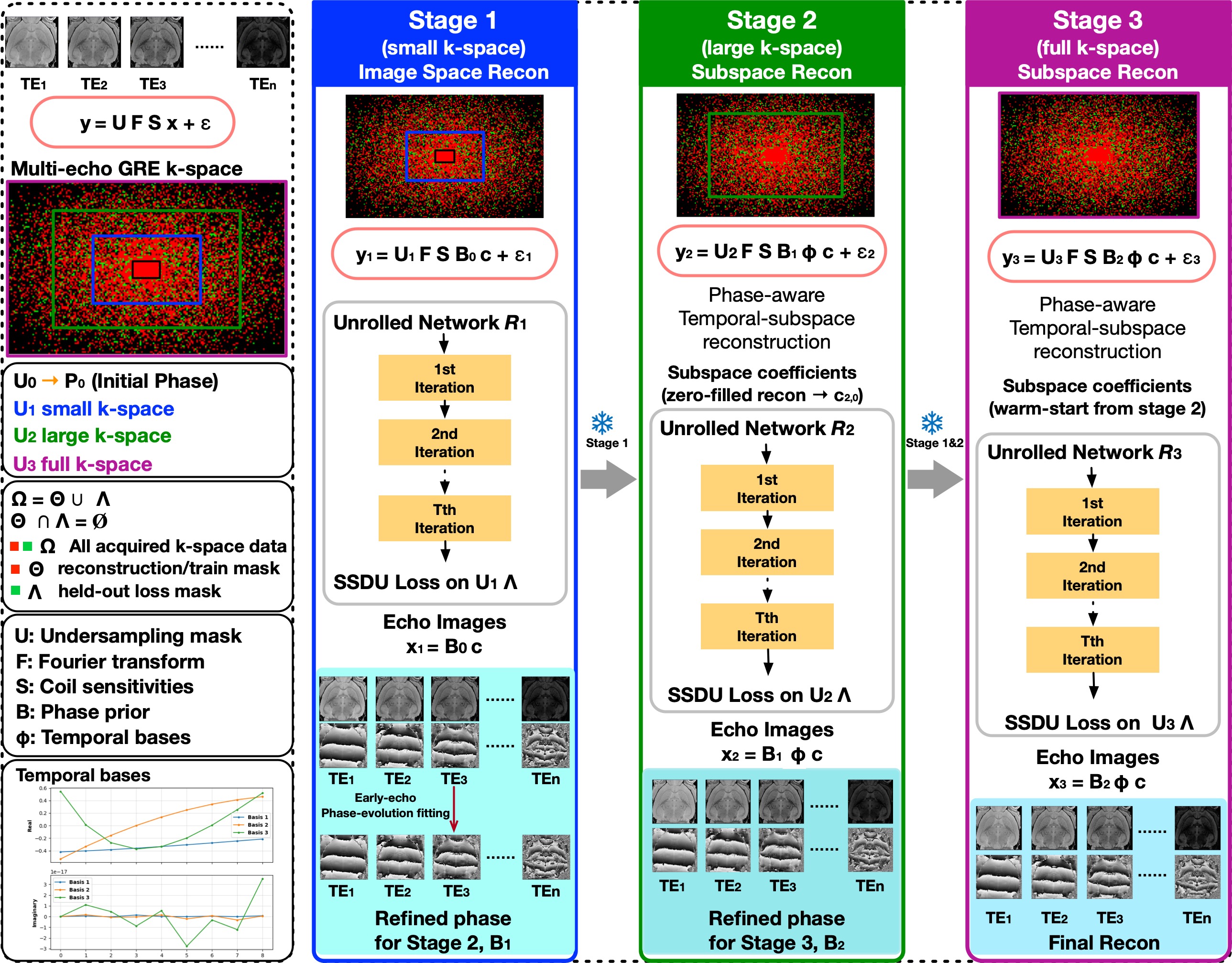}
    \caption{Overview of the proposed multi-stage phase-aware self-supervised reconstruction (MSPR-GRE). During training, the acquired k-space measurements $\Omega$ are partitioned into reconstruction samples $\Theta$ and held-out samples $\Lambda$. Three sequential reconstruction stages progressively expand the k-space support and update the phase estimate. Stage~1 forms its initial phase estimate from low-resolution, coil-combined images reconstructed using only the central k-space samples in its support $\Theta^{(1)}$, and reconstructs the multi-echo images from that support. Early-echo phase-evolution fitting of the Stage~1 reconstruction generates the phase prior for Stage~2, which incorporates additional k-space measurements. Stage~3 extracts the phase directly from the Stage~2 reconstructed complex images and reconstructs from the full reconstruction support. For acquisitions with at least four echoes, Stages~2 and~3 use a relaxation-informed temporal subspace, whereas all stages operate in the image domain for shorter echo trains. The networks are trained sequentially, with preceding stages frozen. Held-out measurements are reserved for the self-supervised loss and excluded from reconstruction, initialization, phase estimation, and data consistency. During inference, all acquired measurements are used for reconstruction.}
    \label{fig:mgre_workflow}
\end{figure}

\subsection{Progressive Phase-Aware Subspace Reconstruction}

MSPR-GRE employs sequential reconstruction networks, denoted by $R_s$, with progressive k-space expansion and inter-stage phase refinement (Figure~\ref{fig:mgre_workflow}). During self-supervised training, the acquired k-space support $\Omega$ is partitioned into disjoint reconstruction and held-out subsets, $\Omega=\Theta\mathbin{\dot\cup}\Lambda$. The reconstruction supports are progressively expanded, satisfying $\Theta^{(1)}\subseteq\Theta^{(2)}\subseteq\cdots\subseteq\Theta^{(S)}=\Theta$, where $S$ denotes the number of reconstruction stages. Specifically, for a three-stage configuration ($S=2$), Stage~1 uses the central $0.5\times0.5$ region of k-space, corresponding to 50\% of the k-space extent along each phase-encoding direction, while Stage~2 expands to the full $1\times1$ k-space extent. For configurations with additional stages, the k-space support is progressively expanded through intermediate regions until the full reconstruction support is reached.

This progressive strategy follows a coarse-to-fine reconstruction approach, analogous to multi-resolution reconstruction. Central k-space primarily contains low-spatial-frequency information that determines the overall image contrast and coarse anatomical structure. Moreover, central k-space is typically densely sampled and contains relatively high signal energy, facilitating more reliable initial reconstruction under undersampled and low-SNR conditions. Since the primary objective of Stage~1 is to provide an initial reconstruction for subsequent phase estimation rather than to recover fine anatomical details, reconstruction from central k-space is sufficient for this purpose. Subsequent stages progressively incorporate higher spatial frequencies to recover finer anatomical structures while refining the phase estimates. This coarse-to-fine strategy enables phase estimation and image reconstruction to be refined progressively as more k-space information becomes available. 

Using a phase-informed subspace formulation, the encoding model for Stage~$s$ is expressed as

\begin{equation}
    \mathbf{y}_{\Theta^{(s)}}
    =\mathbf{U}_{\Theta^{(s)}}\mathbf{F}\mathbf{S}
    \mathbf{B}_{s-1}\boldsymbol{\phi}_s
    \mathbf{c}^{(s)}
    +\boldsymbol{\varepsilon}_{\Theta^{(s)}},
    \label{eq:unified_forward}
\end{equation}

where $\mathbf{B}_{s-1}$ represents spatially varying phase evolution across echoes, $\boldsymbol{\phi}_s$ is the stage-specific temporal basis, and $\mathbf{c}^{(s)}$ contains the reconstruction coefficients. The phase and temporal-basis operators are applied across the multi-echo images.

The reconstruction representation is adapted to the number of echoes. For $N_e\geq4$, Stages~2 and~3 employ a relaxation-informed temporal basis $\boldsymbol{\phi}\in\mathbb{C}^{N_e\times L}$, where $L<N_e$, to reduce the number of unknown coefficient maps. The basis is constructed using singular value decomposition (SVD) of simulated GRE signal evolutions and satisfies $\boldsymbol{\phi}^H\boldsymbol{\phi}=\mathbf{I}_L$. For shorter echo trains, reconstruction is performed directly in the image domain. The stage-specific temporal basis is defined as

\begin{equation}
    \boldsymbol{\phi}_s=
    \begin{cases}
        \mathbf{I}_{N_e}, & s=1\ \text{or}\ N_e\leq3,\\
        \boldsymbol{\phi}, & s\in\{2,3\},\ N_e\geq4.
    \end{cases}
    \label{eq:stage_basis}
\end{equation}

Thus, $\mathbf{c}^{(s)}$ represents phase-demodulated multi-echo images in the image-domain configuration and temporal-subspace coefficient maps otherwise.

All three reconstruction stages approximately solve

\begin{equation}
\begin{aligned}
    \widehat{\mathbf{c}}^{(s)}
    =\underset{\mathbf{c}}{\operatorname{arg\,min}}\quad&
    \left\|
    \mathbf{A}_{\Theta^{(s)}}\mathbf{B}_{s-1}
    \boldsymbol{\phi}_s\mathbf{c}
    -\mathbf{y}_{\Theta^{(s)}}
    \right\|_2^2\\
    &+\mu_s\mathcal{R}_{\theta_s}(\mathbf{c}),
\end{aligned}
\label{eq:unified_recon}
\end{equation}

where $\mathcal{R}_{\theta_s}$ is a learned regularizer and $\mu_s>0$ controls its strength, with the value selected based on the noise level in the acquired k-space data. Each stage uses an unrolled network (Supplementary Figure~\ref{fig:Fig_mGRE_unrollednetwork}) composed of iterative blocks that alternate a convolutional neural network (CNN)-based regularization step with data consistency in complex k-space. The phase operator and temporal basis remain fixed within each reconstruction stage.

\subsubsection{Stage 1: Image-Domain Reconstruction}

Stage~1 jointly reconstructs phase-demodulated multi-echo images using the central k-space samples in $\Theta^{(1)}$. Its initial phase estimate (prior), $\widehat{\Phi}_{\mathrm{LR}}$, is obtained from low-resolution, coil-combined complex images reconstructed using only those same samples. Thus, no measurements outside $\Theta^{(1)}$, including held-out samples in $\Lambda$, contribute to the Stage~1 phase estimate. The initial phase factor is defined as

\begin{equation}
    b_0(\mathbf{r},t_e)
    =\exp\left(i\widehat{\Phi}_{\mathrm{LR}}(\mathbf{r},t_e)\right),
    \label{eq:stage1_phase}
\end{equation}

where the corresponding phase operator $\mathbf{B}_0$ applies elementwise multiplication by $b_0$ across spatial locations and echoes.

Stage~1 is initialized using the phase-demodulated adjoint reconstruction:

\begin{equation}
    \mathbf{c}_0^{(1)}
    =\mathbf{B}_0^H
    \mathbf{A}_{\Theta^{(1)}}^H
    \mathbf{y}_{\Theta^{(1)}}.
    \label{eq:stage1_init}
\end{equation}

The reconstructed complex multi-echo images are subsequently used to estimate the phase evolution for Stage~2.

\subsubsection{Stage 2: Early-Echo Phase-Informed Subspace Reconstruction}

Stage~2 uses the expanded sampling support $\Theta^{(2)}$ and a phase prior derived from the Stage~1 reconstruction. For $N_e\geq3$, a phase-evolution model is fitted to the first $K$ complex images reconstructed by Stage~1 ($3\leq K\leq N_e$), with $K$ chosen to include early echoes whose phase estimates remain reliable. Nonlinear complex-data fitting, adapted from a field-map estimation method used in QSM \cite{liu2013nonlinear}, estimates the phase offset and off-resonance frequency from these echoes. The fitted parameters are used only to synthesize the Stage~2 phase prior across all echo times: phase evolution estimated from the reliable early echoes is extrapolated to later echoes, where the Stage~1 reconstruction may have larger phase errors. The fitted off-resonance frequency is therefore an intermediate parameter, not a quantitative field-map output; the goal is a useful phase prior rather than an accurate field map. For $N_e<3$, there are too few echoes for this fit, so the phase prior is obtained directly from the phase of the Stage~1 reconstructed complex images. No explicit phase unwrapping is performed. The Stage~2 phase factor is

\begin{equation}
b_1(\mathbf{r},t_e)=
\begin{cases}
\displaystyle
\exp\left[
i\left(
\widehat{\phi}_0^{(1)}(\mathbf{r})
+2\pi t_e\widehat{\Delta f}^{(1)}(\mathbf{r})
\right)\right],
& N_e\geq3,\\[8pt]
\displaystyle
\exp\left(
i\arg\left[
\widehat{X}^{(1)}(\mathbf{r},t_e)
\right]\right),
& N_e<3,
\end{cases}
\label{eq:stage2_phase}
\end{equation}

where $\widehat{\phi}_0^{(1)}$ and $\widehat{\Delta f}^{(1)}$ are the fitted phase offset and off-resonance frequency, respectively. They are used only to generate the phase factors $b_1$ at each echo time, not interpreted as quantitative field-map estimates. The phase operator $\mathbf{B}_1$ applies these factors across echoes, including those beyond the fitting range.

To avoid propagating potential Stage-1 artifacts in regions of rapid phase variation, Stage~2 is initialized using a fresh adjoint reconstruction from the expanded sampling support, followed by phase demodulation and temporal-basis projection:

\begin{equation}
    \mathbf{c}_0^{(2)}
    =\boldsymbol{\phi}_2^H\mathbf{B}_1^H
    \mathbf{A}_{\Theta^{(2)}}^H
    \mathbf{y}_{\Theta^{(2)}}.
    \label{eq:stage2_init}
\end{equation}

The resulting coefficient maps are refined using Eq.~\eqref{eq:unified_recon}, with the phase operator $\mathbf{B}_1$ and temporal basis $\boldsymbol{\phi}_2$ held fixed during reconstruction.

\subsubsection{Stage 3: Phase-Refined Subspace Reconstruction}

Stage~3 obtains its phase factor directly from the voxelwise phase of the Stage~2 reconstructed complex images:

\begin{equation}
    b_2(\mathbf{r},t_e)
    =\exp\left(
    i\arg\left[
    \widehat{X}^{(2)}(\mathbf{r},t_e)
    \right]\right),
    \label{eq:stage3_phase}
\end{equation}

The corresponding phase operator $\mathbf{B}_2$ applies the refined phase evolution.

Unlike Stage~2, Stage~3 is initialized directly from the Stage-2 reconstruction. The reconstructed images are demodulated using the refined phase and projected onto the temporal basis:

\begin{equation}
    \mathbf{c}_0^{(3)}
    =\boldsymbol{\phi}_3^H\mathbf{B}_2^H
    \widehat{\mathbf{X}}^{(2)}.
    \label{eq:stage3_init}
\end{equation}

Stage~3 then refines the coefficient maps using the full reconstruction support $\Theta^{(3)}=\Theta$, with the phase operator $\mathbf{B}_2$ derived directly from the Stage~2 reconstruction and the temporal basis $\boldsymbol{\phi}_3$ held fixed.

For all stages, the reconstructed complex multi-echo images are synthesized as

\begin{equation}
    \widehat{\mathbf{X}}^{(s)}
    =\mathbf{B}_{s-1}\boldsymbol{\phi}_s
    \widehat{\mathbf{c}}^{(s)}.
    \label{eq:stage_output}
\end{equation}

For acquisitions with fewer than three echoes, phase-evolution fitting is omitted because the echo train is too short for a stable joint estimate of phase offset and off-resonance; the direct phase estimate is retained.

\subsection{Self-Supervised Training and Inference}

MSPR-GRE employs self-supervised training directly on acquired undersampled k-space measurements without fully sampled reference images \cite{batson2019noise2self,yaman2020ssdu}. At each training iteration, the acquired support $\Omega$ is randomly partitioned into disjoint reconstruction and held-out subsets, $\Omega=\Theta\mathbin{\dot\cup}\Lambda$. The partition is fixed within each forward pass and resampled between iterations. The stage-specific reconstruction supports satisfy $\Theta^{(1)}\subseteq\Theta^{(2)}\subseteq\Theta^{(3)}=\Theta$. Held-out measurements are reserved for loss computation and excluded from reconstruction, phase estimation, initialization, and data consistency.

For each stage $s\in\{1,2,3\}$, the predicted held-out k-space measurements are

\begin{equation}
    \widehat{\mathbf{y}}_{\Lambda}^{(s)}
    =\mathbf{A}_{\Lambda}
    \left(\widehat{\mathbf{X}}^{(s)}\right).
    \label{eq:ssdu_heldout_vectors}
\end{equation}

Each stage is optimized using the normalized SSDU loss \cite{yaman2020ssdu}:

\begin{equation}
\begin{aligned}
    \mathcal{L}_{\mathrm{SSDU}}^{(s)}
    &=
    \frac{
    \left\|\widehat{\mathbf{y}}_{\Lambda}^{(s)}
    -\mathbf{y}_{\Lambda}\right\|_1
    }{
    \left\|\mathbf{y}_{\Lambda}\right\|_1
    }+
    \frac{
    \left\|\widehat{\mathbf{y}}_{\Lambda}^{(s)}
    -\mathbf{y}_{\Lambda}\right\|_2
    }{
    \left\|\mathbf{y}_{\Lambda}\right\|_2
    }.
\end{aligned}
\label{eq:ssdu_loss}
\end{equation}

The loss combines normalized $\ell_1$ and unsquared $\ell_2$ residuals in complex k-space. The same held-out subset is used across stages within each partition.

The networks are trained sequentially, with previously optimized stages frozen. At inference, the SSDU partition is removed, and all acquired measurements are used to recompute stage-specific initializations and phase estimates. The trained networks are applied sequentially to produce the final complex multi-echo reconstruction $\widehat{\mathbf{X}}^{(3)}$ for subsequent quantitative analysis.

\subsection{Experiments}

Experiments compared MSPR-GRE with conventional and self-supervised reconstruction methods using two ex vivo mouse-brain datasets and the in vivo AHEAD human-brain dataset. Both mouse-brain datasets were acquired from a 5xFAD transgenic mouse, a widely used model of Alzheimer's disease (AD) characterized by amyloid-$\beta$ pathology. Qualitative ablations examined phase handling and phase refinement, and a supervised mGRE experiment assessed phase-informed reconstruction.

\subsubsection{Dataset 1: 30-$\mu$m Multi-Echo GRE Imaging of Ex Vivo Mouse Brain at 9.4~T}

All animal experiments were conducted in accordance with protocols approved by the local Institutional Animal Care and Use Committee (IACUC). An ex vivo mouse brain was imaged at $30\,\mu\mathrm{m}$ isotropic resolution using a three-dimensional multi-echo GRE sequence on a 9.4~T Bruker BioSpec MRI system (Bruker BioSpin, Ettlingen, Germany) equipped with a four-channel RF receiver coil. Three separate acquisitions were performed: two with one and three signal averages, respectively, and a third with one signal average and an acceleration factor of $R=3$. These acquisitions were used to evaluate reconstruction performance under different SNR and k-space undersampling conditions. The imaging parameters were as follows: matrix size, $360\times384\times238$; field of view (FOV), $10.8\times11.52\times7.14$ mm$^3$; isotropic resolution, $30\,\mu\mathrm{m}$; flip angle (FA), $50^\circ$; readout scheme, monopolar; first echo time (TE), 3.0~ms; echo spacing, 3.25~ms; number of echoes, 8; repetition time (TR), 185~ms; receiver bandwidth (RBW), 250~kHz; and acquisition time, 4.17~h per average. These datasets were used to evaluate reconstruction quality, phase refinement, and the effects of multiple stages.

Raw multi-coil k-space data were preprocessed using coil-noise prewhitening and ESPIRiT-based coil-sensitivity estimation implemented in the Berkeley Advanced Reconstruction Toolbox (BART) \cite{uecker2014espirit,uecker2015berkeley}. For the three-average acquisition, we reconstructed images using inverse Fourier transformation with coil combination and, separately, using the proposed MSPR-GRE method. For the single-average fully sampled and retrospectively accelerated acquisitions, four methods were compared: conventional inverse Fourier reconstruction with coil combination, phase-aware LLR-regularized subspace reconstruction (abbreviated as LLR; implementation details are provided in the Supplementary Materials), SSDU, and MSPR-GRE. SSDU and MSPR-GRE used the same unrolled network architecture during training (Figure~\ref{fig:Fig_mGRE_unrollednetwork}). This comparison assessed reconstruction performance under reduced signal averaging. The resulting complex multi-echo images were further processed using DECOMPOSE-QSM to generate $T_2^*$ maps, QSM, diamagnetic component susceptibility (DCS), and paramagnetic component susceptibility (PCS) \cite{chen2021decompose}, enabling evaluation of downstream quantitative imaging.

For both SSDU and MSPR-GRE, a one-dimensional inverse Fourier transform was first applied to the 3D k-space data along the readout direction. Because of GPU memory constraints, the networks were trained using a 2.5D strategy with nine adjacent slices as input. Complex-valued inputs were represented by concatenating their real and imaginary components along the channel dimension. For MSPR-GRE, the relaxation-informed temporal basis was constructed from simulated GRE signals spanning residual off-resonance frequencies from $-10$ to $10$~Hz and $T_2^*$ values from $5$ to $150$~ms. The three leading temporal basis functions were retained for subspace reconstruction. In MSPR-GRE, the first four echoes are used for phase-evolution fitting after Stage~1 to estimate the refined phase prior for Stage~2.

To further evaluate reconstruction performance under accelerated acquisition, the single-average k-space data were retrospectively undersampled by a factor of $R=3$ using echo-specific Poisson sampling patterns. Four methods were compared: zero-filled reconstruction, LLR, SSDU, and MSPR-GRE. The three-average reconstruction was used as a higher-SNR reference for quantitative comparison. Reconstruction accuracy and structural fidelity were assessed using normalized root mean square error (NRMSE), peak signal-to-noise ratio (PSNR), and structural similarity index (SSIM).

\subsubsection{Dataset 2: AHEAD Human Brain Dataset at 7~T}

The Amsterdam Ultra-high-field Adult Lifespan (AHEAD) database comprises high-resolution whole-brain structural MRI data acquired using a Philips Achieva 7~T MRI scanner equipped with a 32-channel head-array coil. The sagittal MP2RAGEME sequence combines two inversion-prepared readouts, with four gradient echoes acquired during the second readout to support quantitative relaxation and susceptibility mapping. The acquisition parameters were as follows: TEs of the second readout, 3.0, 11.5, 19.0, and 28.5~ms; TE of the first readout, 3.0~ms; inversion times (TIs), 670 and 3675.4~ms; GRE TRs, 6.2 and 31~ms; sequence TR, 6778~ms; FAs, $4^\circ/4^\circ$; FOV, $205\times205\times164$~mm$^3$; acquired spatial resolution, 0.7~mm isotropic; acquisition matrix, $292\times290\times234$; SENSE acceleration factor, $R=2$; and turbo field echo (TFE) factor, 150 \cite{alkemade2020ahead}. The dataset was used to evaluate reconstruction across acceleration factors and quantitative mapping.

Fifteen AHEAD datasets with minimal residual motion artifacts were selected for evaluation. Across all experiments using AHEAD, 10 datasets were used for training and five independent datasets were reserved for testing. Each dataset contained multi-coil k-space measurements and corresponding coil-sensitivity maps. The publicly released data had undergone preprocessing, including motion correction and image-domain defacing. Although the original acquisitions employed parallel imaging acceleration, the released k-space data were provided on a complete Cartesian grid following preprocessing. We therefore treated the provided k-space data as fully sampled reference data for the purposes of retrospective undersampling and reconstruction evaluation, rather than as fully sampled raw acquisitions.

To assess reconstruction performance across acceleration levels, the provided k-space data were retrospectively undersampled using sampling patterns with nominal acceleration factors of $R=4$, $8$, and $12$, defined relative to the released Cartesian k-space grid. We compared three methods: $\ell_1$-wavelet regularization, SSDU, and MSPR-GRE. SSDU and MSPR-GRE followed the same training strategy described for Dataset~1. The MSPR-GRE model used three temporal basis functions and a three-stage reconstruction. Reconstructed complex multi-echo images were used to generate $T_2^*$ and QSM maps, enabling evaluation of anatomical image quality and downstream quantitative mapping. NRMSE, PSNR, and SSIM were calculated relative to reference images reconstructed from the provided baseline k-space data.


\subsubsection{Dataset 3: 18-$\mu$m Dual-Echo GRE Imaging of Ex Vivo Mouse Brain at 9.4~T}

An ex vivo 5xFAD mouse brain was imaged at $18\,\mu\mathrm{m}$ isotropic resolution using a 3D dual-echo GRE sequence on a 9.4~T Bruker BioSpec MRI system with a four-channel RF receiver coil. Acquisition parameters were: matrix size, $850\times640\times400$; FOV, $15.3\times11.52\times7.2$ mm$^3$; FA, $50^\circ$; TEs, 8.00 and 19.92~ms; TR, 105~ms; RBW, 91~kHz; NEX, 1; and acquisition time, approximately 5.3~h per scan. Four independent, fully sampled single-average acquisitions were performed. This dataset was used to evaluate the applicability of MSPR-GRE to shorter echo trains and very high spatial resolution.

Raw k-space data were preprocessed using coil-noise prewhitening and ESPIRiT coil-sensitivity estimation, as described for Dataset~1. Each acquisition was independently reconstructed using inverse Fourier transformation with coil combination and MSPR-GRE. Complex images from the four fully sampled acquisitions were globally phase-aligned and averaged echo-wise to generate a higher-SNR reference.

For single-average evaluation, one fully sampled acquisition was reconstructed using inverse Fourier reconstruction, total variation (TV) regularization, SSDU, and MSPR-GRE. The same acquisition was also retrospectively undersampled at $R=3$ using an echo-specific Poisson sampling pattern. For the undersampled data, we compared zero-filled reconstruction, TV-regularized reconstruction, SSDU, and MSPR-GRE. Reconstructed complex images were used to derive $T_2^*$ and QSM. Reconstruction quality was assessed using NRMSE, PSNR, and SSIM relative to the four-scan averaged reference.

\subsection{Ablation and Supervised Experiments}

We conducted ablation and supervised experiments to evaluate the contributions of phase-aware reconstruction and multi-stage reconstruction:

\begin{itemize}
    \item \textbf{Phase-aware reconstruction and multi-stage reconstruction (Dataset~1).}
    We compared conventional SSDU, single-stage SSDU with phase-aware reconstruction, and the full MSPR-GRE framework to assess the benefits of phase-aware reconstruction and progressive multi-stage reconstruction. In addition, we compared MSPR-GRE models with one, two, three, or four reconstruction stages to evaluate the effect of the number of reconstruction stages.

    \item \textbf{Early-echo phase-evolution fitting and phase refinement (Dataset~1).}
    We compared MSPR-GRE with a variant that omitted the Stage~1 early-echo phase-evolution fit for Stage~2 initialization. Phase estimates and reconstructed images across stages were also examined to assess late-echo phase consistency and anatomical detail.

    \item \textbf{Supervised reconstruction (Dataset~2).}
    To evaluate phase-aware reconstruction in a supervised learning setting, we trained unrolled reconstruction networks without SSDU partitioning, using images reconstructed from all available k-space data as reference targets and an image-domain $L_1$ loss for supervision. Three network configurations were evaluated: vanilla, a single-stage unrolled network without phase-aware reconstruction; vanilla+PA, a single-stage unrolled network incorporating phase-aware reconstruction; and MSPR, a three-stage unrolled network with phase-aware reconstruction. These configurations enabled assessment of the individual contributions of phase-aware reconstruction and the multi-stage reconstruction strategy to supervised reconstruction performance.

\end{itemize}

\section{Results}

\subsection{Reconstruction Comparisons Across Datasets}

\subsubsection{Dataset 1: 30-$\mu$m Multi-Echo GRE Imaging of Ex Vivo Mouse Brain at 9.4 T}

\begin{figure}[!t]
    \centering
    \includegraphics[width=\textwidth]{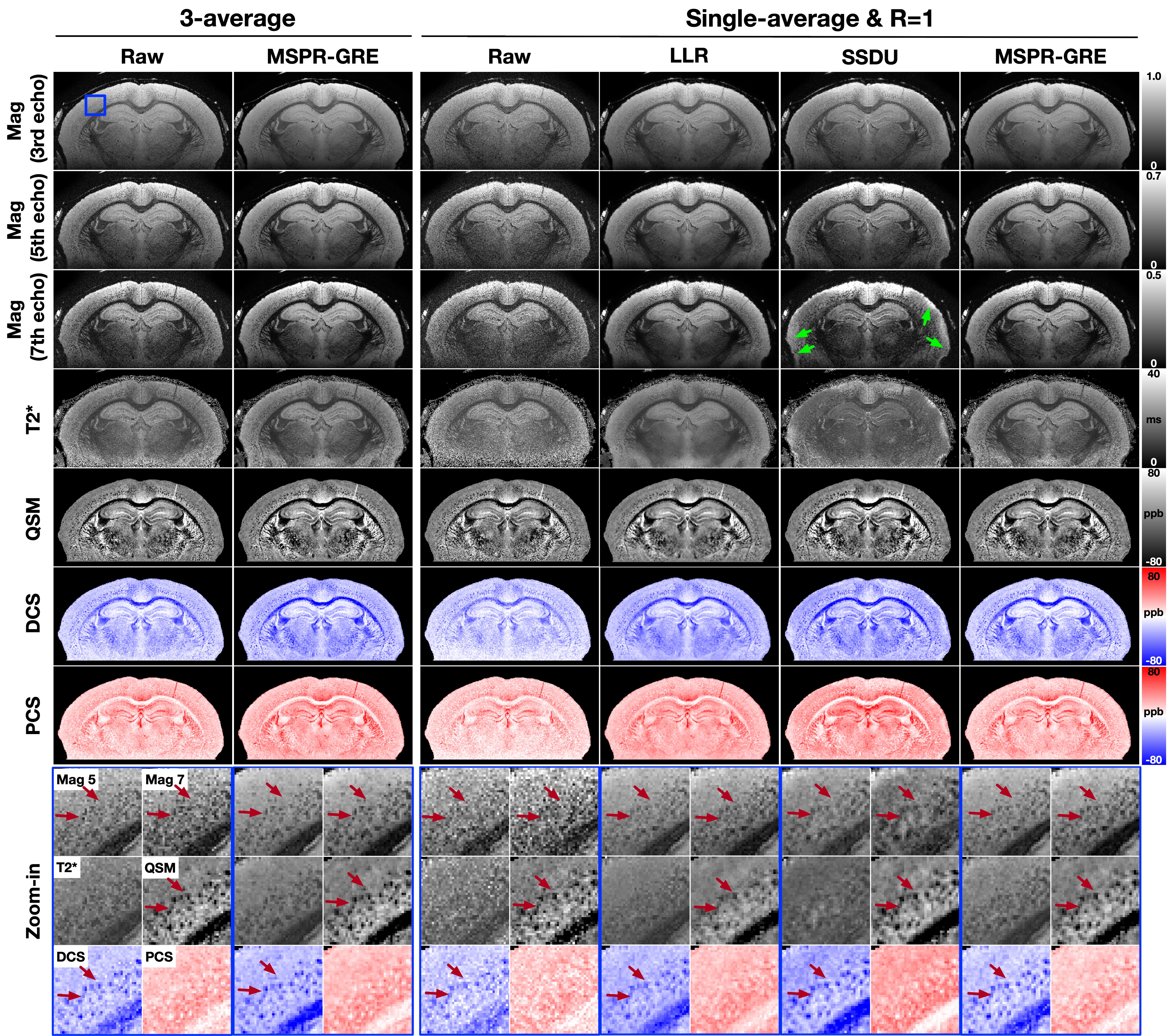}
    \caption{Comparison of reconstruction methods for fully sampled 30-$\mu$m multi-echo GRE data. The three-average reconstruction provides higher SNR than the single-average reconstruction but retains residual noise. Green arrows highlight artifacts in SSDU reconstructions, while magnified views illustrate the preservation of fine anatomical structures, including hypointense cortical plaques (red arrows).}
    \label{fig:dataset1_reconstruction}
\end{figure}

Figure~\ref{fig:dataset1_reconstruction} compares reconstruction methods for fully sampled 30-$\mu$m multi-echo GRE data. The three-average reconstruction exhibited higher SNR than the single-average reconstruction but retained residual noise. LLR produced magnitude images with good anatomical detail; however, the derived $T_2^*$ maps appeared overly smoothed, potentially due to the relatively strong regularization required for noise suppression. Compared with SSDU, MSPR-GRE reduced visible artifacts and better preserved fine anatomical structures, including hypointense features consistent with amyloid-$\beta$ plaques in the cortical QSM and DCS images (red arrows).

\begin{figure}[!t]
    \centering
    \includegraphics[width=\textwidth]{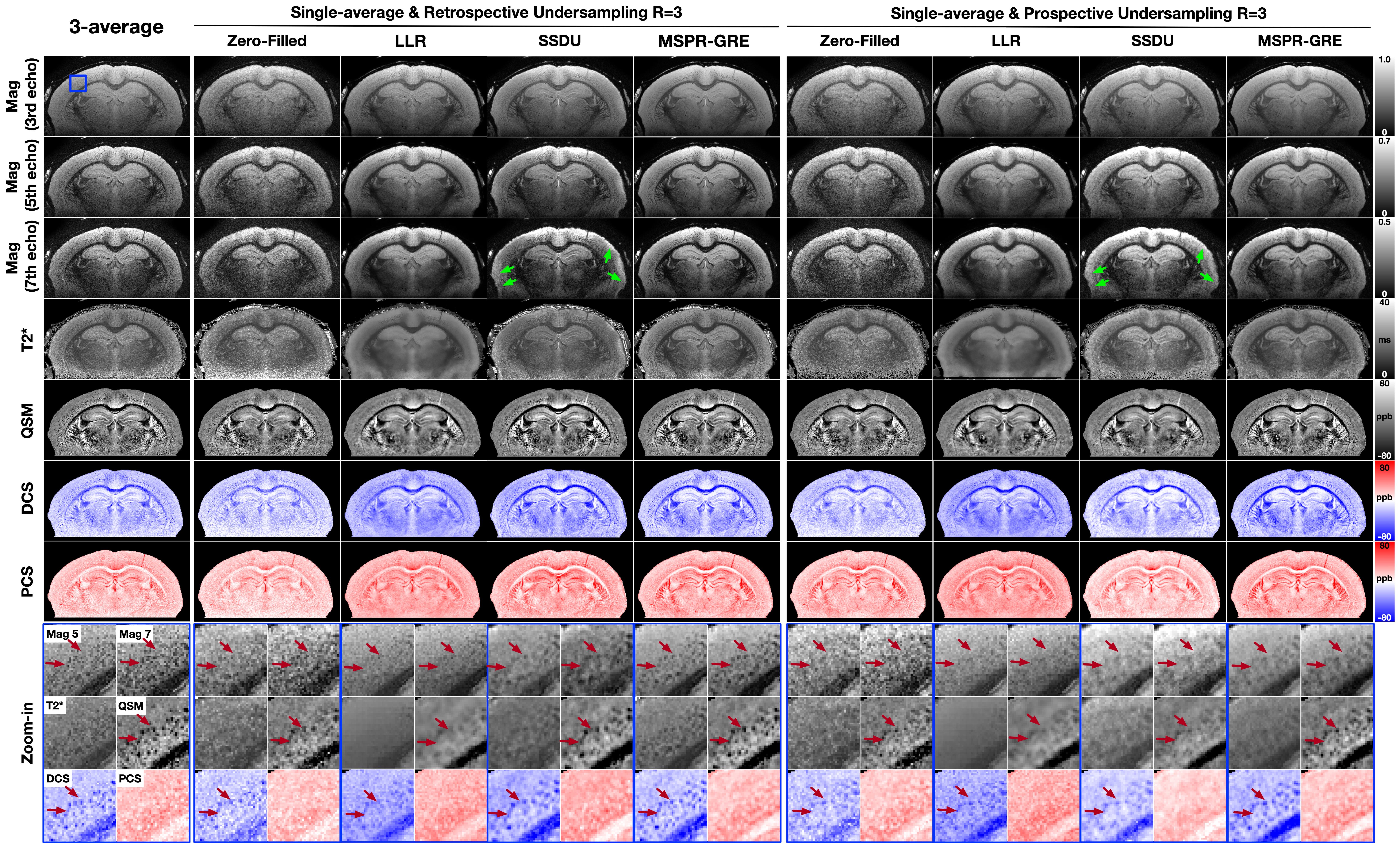}
    \caption{Comparison of reconstruction methods for 30-$\mu$m multi-echo GRE data with prospective and retrospective undersampling at an acceleration factor of $R=3$. Zero-filled reconstruction exhibits substantial noise, which also affects the derived $T_2^*$ and susceptibility maps. SSDU reduces noise but retains visible artifacts in regions of rapid phase variation (green arrows). In contrast, MSPR-GRE provides improved noise suppression and preservation of fine anatomical details, with fewer visible artifacts in both the reconstructed images and quantitative maps. Magnified views highlight the improved visibility of small hypointense cortical plaques in MSPR-GRE reconstructions. Compared to prospective undersampling, the retrospectively undersampled reconstructions preserve anatomical details more effectively.}
    \label{fig:Fig_mGRE_30um_acc3}
\end{figure}

Figure~\ref{fig:Fig_mGRE_30um_acc3} compares reconstruction methods under retrospective and prospective undersampling at $R=3$. Under both conditions, MSPR-GRE reduced visible artifacts and better preserved fine anatomical structures than SSDU. Although LLR provided reasonable magnitude reconstructions, the corresponding $T_2^*$ maps exhibited excessive smoothing, suggesting a trade-off between noise suppression and preservation of fine structural details. In contrast, MSPR-GRE better preserved anatomical contrast in both the magnitude images and derived quantitative maps.

Table~\ref{tab:dataset1_metrics} summarizes quantitative reconstruction performance using the three-average conventional reconstruction as the reference. MSPR-GRE achieved the lowest magnitude NRMSE at both $R=1$ and $R=3$. Compared with SSDU, NRMSE decreased from 16.69 to 14.27 for the third echo and from 24.65 to 20.46 for the fifth echo at $R=1$, with similar improvements at $R=3$ (17.25 to 14.30 and 25.12 to 20.63, respectively). For QSM, MSPR-GRE achieved the highest SSIM at both acceleration factors (0.981 and 0.980), although LLR yielded slightly lower NRMSE at $R=1$. Quantitative $T_2^*$ metrics were excluded due to substantial noise in the reference maps. The reported metrics should be interpreted with caution given the residual noise in the reference reconstruction.

\begin{table}[t]
    \centering
    \caption{Quantitative reconstruction performance for the 30-$\mu$m multi-echo GRE dataset under fully sampled ($R=1$) and retrospectively undersampled ($R=3$) conditions, using the three-average reconstruction as the reference. Bold indicates the best result among the evaluated methods for each output, acceleration factor, and metric (lowest NRMSE; highest PSNR and SSIM).}
    \label{tab:dataset1_metrics}
    \small
    \setlength{\tabcolsep}{4pt}
    \renewcommand{\arraystretch}{1.05}
    \begin{tabular*}{\textwidth}{@{\extracolsep{\fill}}llrrrrrr@{}}
        \toprule
        \textbf{Output} & \textbf{Method} & \multicolumn{3}{c}{$R=1$} & \multicolumn{3}{c}{$R=3$} \\
        \cmidrule(lr){3-5}\cmidrule(lr){6-8}
        & & \textbf{NRMSE ($\%$)$\downarrow$} & \textbf{PSNR $\uparrow$} & \textbf{SSIM $\uparrow$} & \textbf{NRMSE ($\%$)$\downarrow$} & \textbf{PSNR $\uparrow$} & \textbf{SSIM $\uparrow$} \\
        \midrule
        Magnitude, 3rd echo 
         & Raw & 19.77 & 28.51 & 0.966 & --- & --- & --- \\
         & Zero-Filled & --- & --- & --- & 19.24 & 28.75 & 0.951 \\
         & LLR & 15.07 & 30.87 & 0.978 & 16.26 & 30.21 & 0.973 \\
         & SSDU & 16.69 & 29.98 & 0.964 & 17.25 & 29.70 & 0.960 \\
         & MSPR-GRE & \textbf{14.27} & \textbf{31.34} & \textbf{0.978} & \textbf{14.30} & \textbf{31.32} & \textbf{0.978} \\

        \addlinespace
        Magnitude, 5th echo 
         & Raw & 28.43 & 26.30 & 0.957 & --- & --- & --- \\
         & Zero-Filled & --- & --- & --- & 26.45 & 26.92 & 0.944 \\
         & LLR & 20.70 & 29.05 & 0.972 & 21.23 & 28.83 & 0.966 \\
         & SSDU & 24.65 & 27.54 & 0.947 & 25.12 & 27.37 & 0.942 \\
         & MSPR-GRE & \textbf{20.46} & \textbf{29.15} & \textbf{0.972} & \textbf{20.63} & \textbf{29.08} & \textbf{0.971} \\

        \addlinespace
        QSM 
         & Raw & 88.26 & 30.82 & 0.975 & --- & --- & --- \\
         & Zero-Filled & --- & --- & --- & 79.19 & 31.76 & 0.968 \\
         & LLR & \textbf{66.91} & \textbf{33.23} & 0.978 & 71.60 & 32.64 & 0.969 \\
         & SSDU & 70.35 & 32.79 & 0.977 & 70.00 & 32.83 & 0.975 \\
         & MSPR-GRE & 67.50 & 33.15 & \textbf{0.981} & \textbf{68.25} & \textbf{33.05} & \textbf{0.980} \\
        \bottomrule
    \end{tabular*}
\end{table}

\subsubsection{Dataset 2: AHEAD Human Brain Dataset at 7 T}

\begin{figure}[t]
    \centering
    \includegraphics[width=\textwidth]{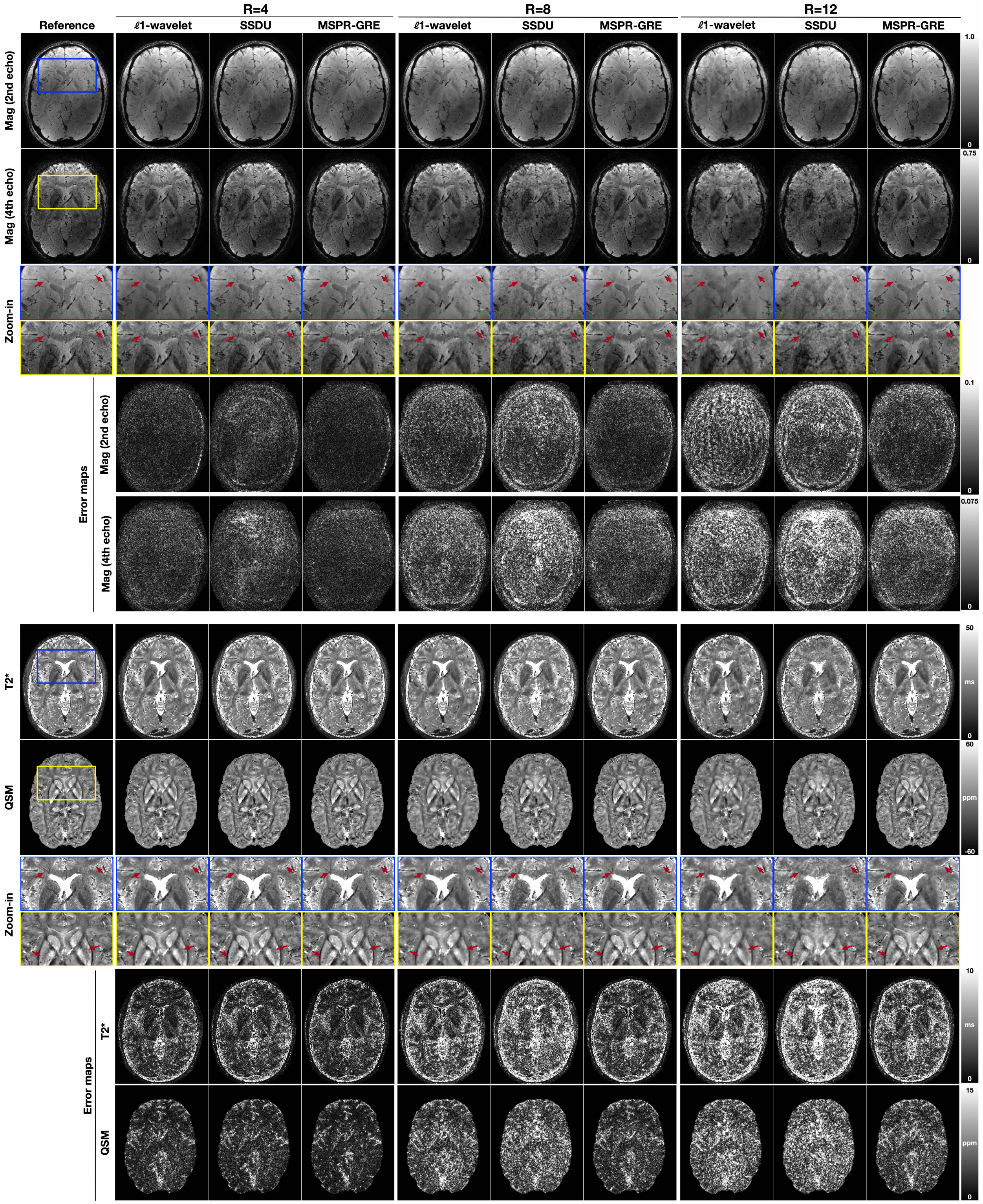}
    \caption{Magnitude reconstructions of the second and fourth echoes and quantitative maps from a representative AHEAD dataset. Reference images and reconstructions obtained using $\ell_1$-wavelet regularization, SSDU, and MSPR-GRE are shown at acceleration factors of $R=4$, $R=8$, and $R=12$. Blue and yellow boxes indicate regions displayed in the magnified views, with red arrows highlighting fine anatomical structures. Error maps show the absolute errors relative to the reference data. MSPR-GRE consistently preserves anatomical details and achieves the lowest reconstruction errors across all acceleration factors. For both $T_2^*$ and QSM maps, MSPR-GRE provides superior image quality with the smallest errors.}
    \label{fig:ahead_mspr_gre}
\end{figure}

Figure~\ref{fig:ahead_mspr_gre} compares magnitude reconstructions and quantitative maps from $\ell_1$-wavelet regularization, SSDU, and MSPR-GRE at different acceleration factors using a representative AHEAD dataset. Reconstruction errors increased with acceleration for all methods, particularly at later echoes with reduced SNR due to $T_2^*$ decay. While SSDU exhibited substantial blurring and artifacts at $R=8$ and $R=12$, MSPR-GRE better preserved fine anatomical structures and reduced residual errors. These improvements were consistent with the quantitative results across five test datasets (Table~\ref{tab:ahead_metrics}). At $R=12$, MSPR-GRE reduced fourth-echo magnitude NRMSE to $12.09 \pm 0.60\%$, compared with $16.67 \pm 0.76\%$ for $\ell_1$-wavelet regularization and $21.55 \pm 1.12\%$ for SSDU. Similarly, second-echo NRMSE decreased to $7.16 \pm 0.42\%$, compared with $12.78 \pm 0.62\%$ for SSDU.

MSPR-GRE also improved downstream quantitative mapping, particularly at higher acceleration factors. At $R=12$, $T_2^*$ NRMSE decreased from $74.11 \pm 1.57\%$ with SSDU to $52.00 \pm 2.64\%$ with MSPR-GRE, while QSM NRMSE decreased from $61.98 \pm 2.85\%$ to $50.10 \pm 2.06\%$. Across the evaluated acceleration factors, MSPR-GRE generally achieved the lowest NRMSE and highest PSNR and SSIM for magnitude images and quantitative maps, with the exception of slightly better $T_2^*$ NRMSE and PSNR from $\ell_1$-wavelet regularization at $R=4$. Overall, these results demonstrate improved reconstruction accuracy and quantitative-map fidelity, with more pronounced benefits at higher acceleration factors.

\begin{table}[t]
    \centering
    \caption{Quantitative reconstruction performance on the AHEAD dataset at retrospective undersampling rates of 4, 8, and 12. Values are mean $\pm$ standard deviation across five datasets. NRMSE is reported as a percentage (NRMSE $\times 100$). Bold indicates the best mean result for each output and acceleration factor (lowest NRMSE; highest PSNR and SSIM).}
    \label{tab:ahead_metrics}
    \small
    \setlength{\tabcolsep}{4pt}
    \renewcommand{\arraystretch}{1.05}
    \resizebox{\textwidth}{!}{%
    \begin{tabular}{@{}llrrrrrrrrr@{}}
        \toprule
        \textbf{Output} & \textbf{Method}
        & \multicolumn{3}{c}{$R=4$}
        & \multicolumn{3}{c}{$R=8$}
        & \multicolumn{3}{c}{$R=12$} \\
        \cmidrule(lr){3-5}\cmidrule(lr){6-8}\cmidrule(lr){9-11}
        & & \textbf{NRMSE ($\%$)$\downarrow$} & \textbf{PSNR $\uparrow$} & \textbf{SSIM $\uparrow$}
          & \textbf{NRMSE ($\%$)$\downarrow$} & \textbf{PSNR $\uparrow$} & \textbf{SSIM $\uparrow$}
          & \textbf{NRMSE ($\%$)$\downarrow$} & \textbf{PSNR $\uparrow$} & \textbf{SSIM $\uparrow$}\\
        \midrule

        Magnitude, 2nd echo
         & $\ell_1$-wavelet & 
         $4.39 \pm 0.53$ & $ 39.94 \pm 1.20$ & $0.9585 \pm 0.0080$ & 
         $7.07 \pm 0.55$ & $35.78 \pm 0.94$ & $0.9079 \pm 0.0138$ & 
         $11.31 \pm 0.37$ & $31.68 \pm 0.86$ & $0.8360 \pm 0.0191$ \\

         & SSDU & 
         $5.79 \pm 0.26$ & $37.49 \pm 0.84$ & $0.9510 \pm 0.0072$ & 
         $ 11.03 \pm 0.39$ & $31.89 \pm 0.84$ & $0.8458 \pm 0.0161$ & 
         $ 12.78 \pm 0.62$ & $30.63 \pm 0.88$ & $0.8179 \pm 0.0189$ \\

         & MSPR-GRE & 
         $\mathbf{4.36} \pm 0.21$ & $\mathbf{39.96} \pm 0.73$ & $\mathbf{0.9840} \pm 0.0127$ & 
         $\mathbf{5.80} \pm 0.43$ & $\mathbf{37.49} \pm 1.10$ & $\mathbf{0.9376} \pm 0.0111$ & 
         $\mathbf{7.16} \pm 0.42$ & $\mathbf{35.66} \pm 1.01$ & $\mathbf{0.9154} \pm 0.0142$ \\
        \addlinespace

        Magnitude, 4th echo
         & $\ell_1$-wavelet & 
         $6.66 \pm 1.02$ & $38.90 \pm 1.63$ & $0.9527 \pm 0.0152$ & 
         $11.38 \pm 1.28$ & $34.22 \pm 1.37$ & $0.8870 \pm 0.0287$ & 
         $16.67 \pm 0.76$ & $30.88 \pm 0.86$ & $0.8167 \pm 0.0221$ \\

         & SSDU & 
         $9.38 \pm 0.40$ & $35.86 \pm 0.80$ & $0.9406 \pm 0.0080$ & 
         $19.39 \pm 0.92$ & $29.56 \pm 0.82$ & $0.7955 \pm 0.0170$ & 
         $21.55 \pm 1.12$ & $28.64 \pm 0.88$ & $0.7633 \pm 0.0203$ \\

         & MSPR-GRE & 
         $\mathbf{5.38} \pm 0.23$ & $\mathbf{39.51} \pm 0.54$ & $\mathbf{0.9625} \pm 0.0142$ & 
         $\mathbf{9.70} \pm 0.61$ & $\mathbf{35.58} \pm 0.87$ & $\mathbf{0.9170} \pm 0.0127$ & $\mathbf{12.09} \pm 0.60$ & $\mathbf{33.66} \pm 0.81$ & $\mathbf{0.8840} \pm 0.0155$ \\
        \addlinespace

        $T_2^*$
         & $\ell_1$-wavelet & 
         $33.32 \pm 4.39$ & $24.37 \pm 1.60$ & $0.9241 \pm 0.0284$ & 
         $47.67 \pm 4.66$ & $21.23 \pm 1.57$ & $0.8412 \pm 0.0530$ & 
         $59.44 \pm 4.61$ & $19.31 \pm 1.53$ & $0.7627 \pm 0.0588$ \\

         & SSDU & 
         $38.37 \pm 2.75$ & $23.11 \pm 0.95$ & $0.9067 \pm 0.0166$ & 
         $56.98 \pm 2.49$ & $19.66 \pm 0.82$ & $0.7898 \pm 0.0261$ & 
         $74.11 \pm 1.57$ & $17.37 \pm 0.81$ & $0.6786 \pm 0.0356$ \\

         & MSPR-GRE & 
         $\mathbf{33.24} \pm 2.79$ & $\mathbf{24.45} \pm 1.01$ & $\mathbf{0.9393} \pm 0.0141$ & 
         $\mathbf{43.93} \pm 2.69$ & $\mathbf{21.92} \pm 0.92$ & $\mathbf{0.8812} \pm 0.0192$ & $\mathbf{52.00} \pm 2.64$ & $\mathbf{20.46} \pm 0.89$ & $\mathbf{0.8282} \pm 0.0231$ \\
        \addlinespace

        QSM
         & $\ell_1$-wavelet & 
         $35.33 \pm 2.22$ & $ 39.13 \pm 0.81$ & $0.9417 \pm 0.0109$ & 
         $44.75 \pm 4.26$ & $37.10 \pm 0.89$ & $0.8831 \pm 0.0276$ & 
         $53.15 \pm 4.26$ & $35.59 \pm 0.87$ & $0.8107 \pm 0.0398$ \\

         & SSDU & 
         $37.80 \pm 1.54$ & $38.54 \pm 0.85$ & $0.9301 \pm 0.0099$ & 
         $57.50 \pm 2.69$ & $34.90 \pm 0.93$ & $0.7879 \pm 0.0306$ & 
         $61.98 \pm 2.85$ & $34.24 \pm 0.93$ & $0.7409 \pm 0.0348$ \\

         & MSPR-GRE & 
         $\mathbf{33.38} \pm 1.20$ & $\mathbf{39.51} \pm 0.82$ & $\mathbf{0.9553} \pm 0.0064$ & 
         $\mathbf{43.47} \pm 1.42$ & $\mathbf{37.32} \pm 0.78$ & $\mathbf{0.9075} \pm 0.0113$ 
         & $\mathbf{50.10} \pm 2.06$ & $\mathbf{36.09} \pm 0.87$ & $\mathbf{0.8632} \pm 0.0192$ \\
        \bottomrule
    \end{tabular}%
    }
\end{table}

\subsubsection{Dataset 3: 18-$\mu$m Dual-Echo GRE Imaging of Ex Vivo Mouse Brain at 9.4~T}

\begin{figure}[t]
    \centering
    \includegraphics[width=\textwidth]{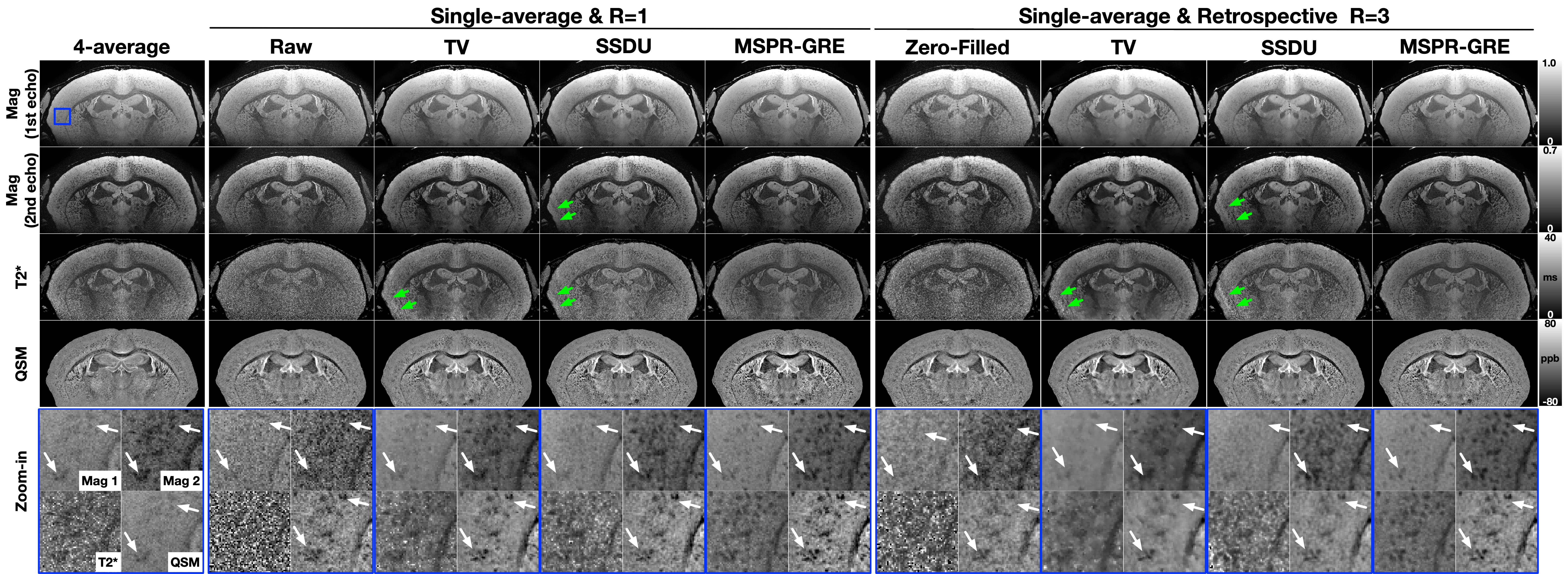}
    \caption{Comparison of reconstruction methods for 18-$\mu$m dual-echo GRE imaging. MSPR-GRE employs a three-stage image-domain reconstruction for this dual-echo acquisition. The four-average reference provides higher SNR than the single-average reconstruction but retains residual noise. Reconstructions from single-average data are compared at $R=1$ (fully sampled) and $R=3$ (retrospectively undersampled). Green arrows highlight artifacts in SSDU reconstructions, while magnified views demonstrate improved noise suppression and preservation of fine anatomical structures with MSPR-GRE, including hypointense cortical plaques (white arrows).}
    \label{fig:dataset2_reconstruction}
\end{figure}

Figure~\ref{fig:dataset2_reconstruction} compares the reconstruction methods on 18-$\mu$m dual-echo GRE data. The four-average reference exhibited higher SNR than the single-average reconstructions but retained residual noise. TV regularization reduced noise compared with conventional single-average reconstruction, while both SSDU and MSPR-GRE provided effective noise suppression. However, SSDU exhibited visible artifacts in regions of rapid phase variation (green arrows) and reduced fine anatomical detail compared with MSPR-GRE. At both $R=1$ and $R=3$, MSPR-GRE reduced these artifacts while better preserving fine structures, including hypointense plaques in the cortex.

Additional reconstruction results and quantitative metrics are provided in the Supplementary Materials. However, the four-average reference images exhibit residual noise, potentially arising from system drift and gradient heating occurring during the prolonged acquisition. These imperfections may compromise the reliability of reference-based metrics and limit their ability to accurately reflect the true reconstruction quality. Therefore, these quantitative comparisons should be interpreted with caution. Metrics for $T_2^*$ maps were not reported because the four-average reference $T_2^*$ map contained substantial noise, making it unsuitable for reliable quantitative evaluation.

\subsection{Ablation and Supervised Experiments}

\begin{figure}[t]
    \centering
    \includegraphics[width=\textwidth]{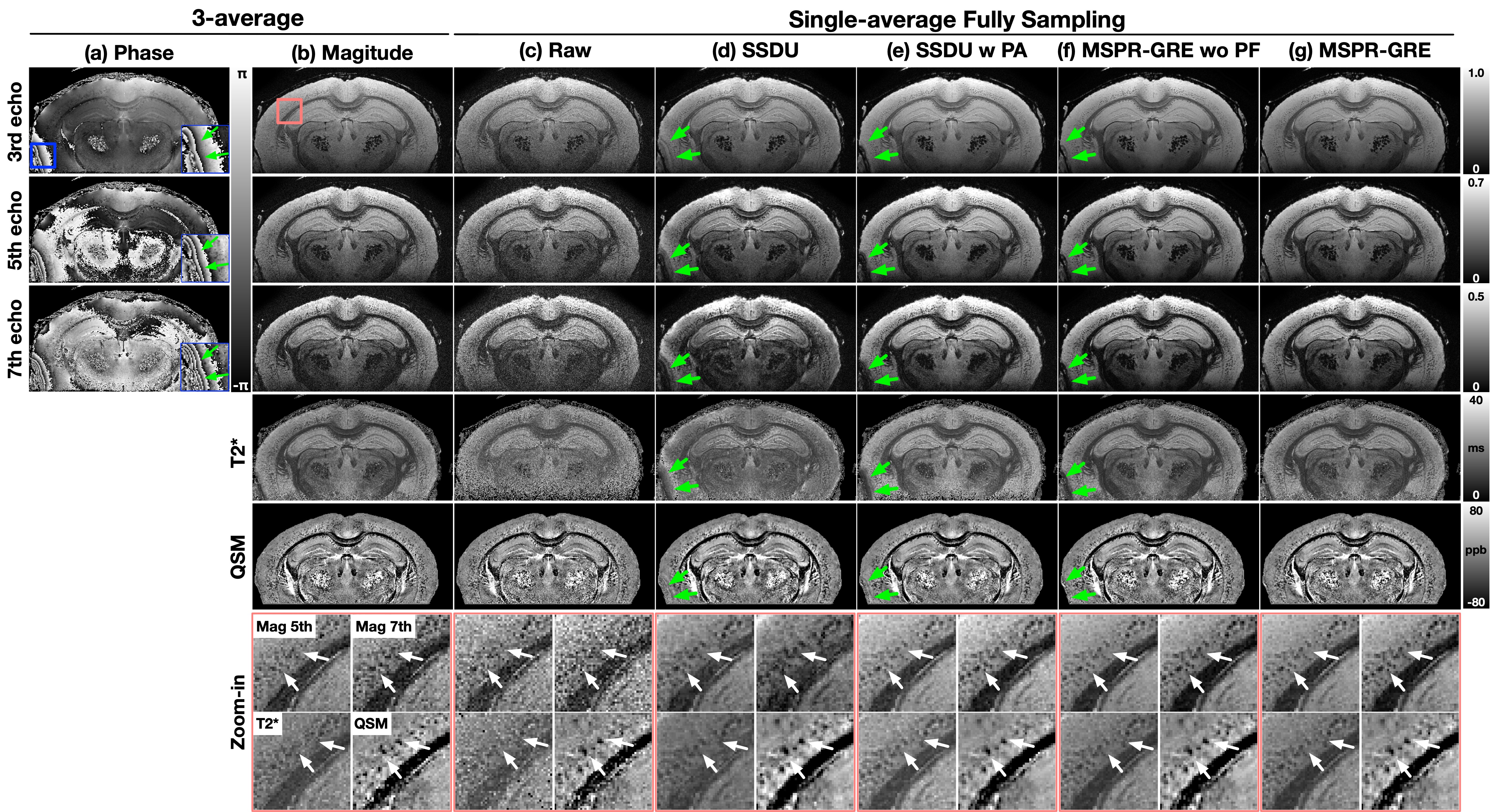}
    \caption{Ablation of phase-aware reconstruction and early-echo phase-evolution fitting. (d) Conventional SSDU without a phase prior, with prominent artifacts, particularly at the fifth and seventh echoes. (e) Phase-aware single-stage SSDU, the single-stage counterpart of MSPR-GRE, with fewer artifacts but residual errors in regions of rapid phase variation. (f) MSPR-GRE without the Stage~1 early-echo phase-evolution fit used to initialize the Stage~2 phase prior. (g) Complete three-stage MSPR-GRE. Green arrows mark artifacts in regions of large off-resonance; white arrows highlight differences in fine anatomical detail.}
    \label{fig:ablation_phase_regularization}
\end{figure}

Figure~\ref{fig:ablation_phase_regularization} compares conventional SSDU, phase-aware reconstruction, and ablations of MSPR-GRE. Conventional SSDU (d) exhibits pronounced artifacts, particularly at the fifth and seventh echoes and in regions of rapid phase variation. Adding phase-aware reconstruction (e), the single-stage phase-aware counterpart of MSPR-GRE, visually improves the reconstruction and reduces these artifacts, although residual errors remain in rapidly varying phase regions. The complete three-stage MSPR-GRE model (g) shows the fewest visible artifacts in these regions. When the Stage~1 early-echo phase-evolution fit used to initialize the Stage~2 phase prior is removed (f), artifacts persist and fine anatomical details differ from those in the complete reconstruction. These qualitative comparisons suggest that both phase-aware reconstruction and early-echo phase fitting contribute to the observed improvement.

\begin{figure}[t]
    \centering
    \includegraphics[width=\textwidth,height=0.85\textheight,keepaspectratio]{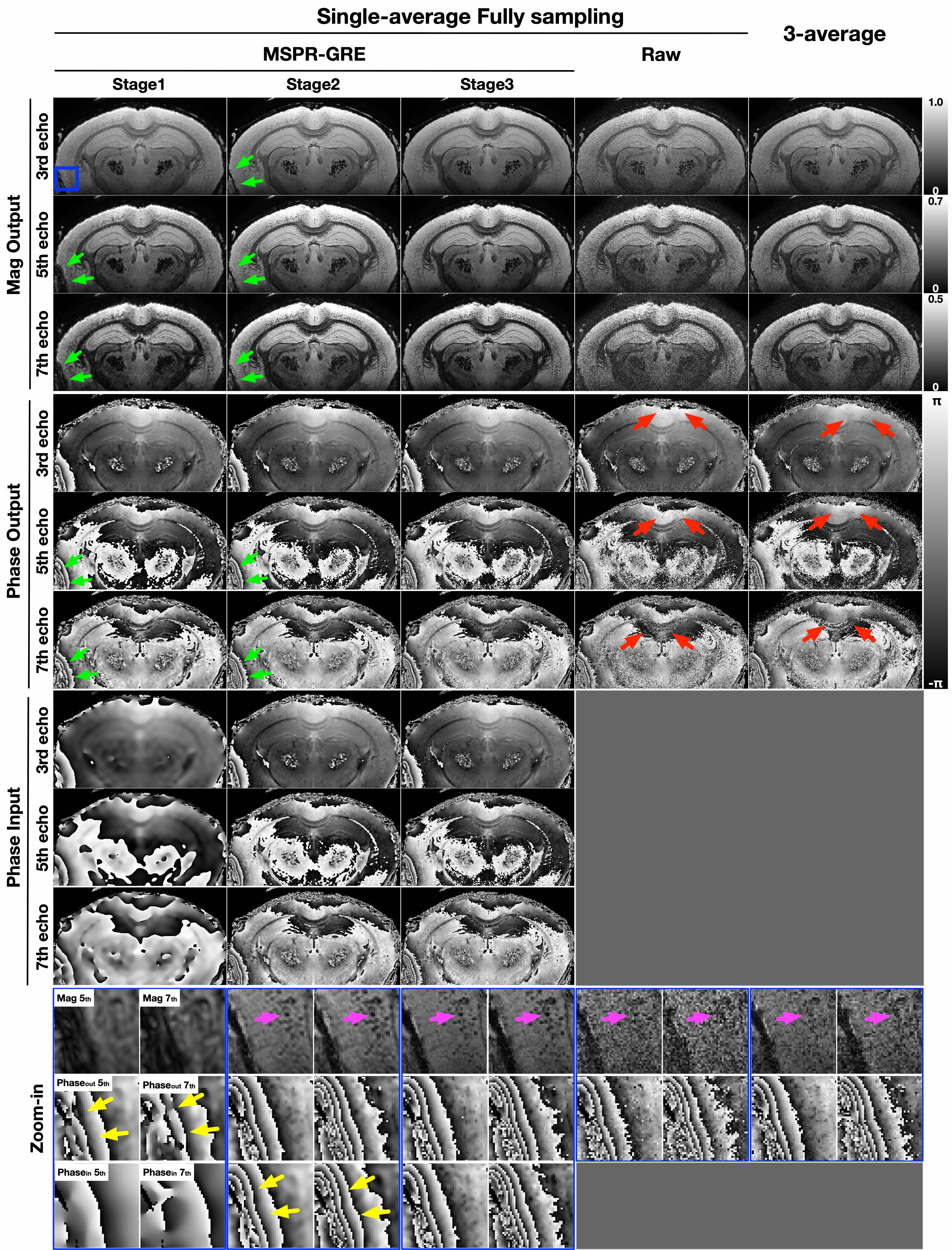}
    \caption{Stage-wise phase estimation and reconstruction in MSPR-GRE. (a) Phase inputs and corresponding magnitude and phase reconstructions across the three stages. Stage~1 uses a low-resolution phase estimate derived from the central k-space samples in $\Theta^{(1)}$, with residual phase errors and magnitude artifacts most pronounced at the fifth and seventh echoes. Stage~2 uses early-echo phase-evolution fitting to generate a phase prior, while Stage~3 takes the phase directly from the Stage~2 reconstructed complex images. Yellow arrows highlight improved phase estimates at the fifth and seventh echoes, and purple arrows indicate enhanced anatomical detail across stages. Residual noise is visible in the three-average reference. Red arrows indicate phase differences between the single-average and three-average acquisitions, potentially associated with scanner drift.}
    \label{fig:fig_mgre_phasechanges}
\end{figure}

Figure~\ref{fig:fig_mgre_phasechanges} shows the phase estimates and corresponding magnitude and phase reconstructions at each MSPR-GRE stage. Stage~1 uses a low-resolution phase estimate derived from central k-space. Residual phase errors and associated magnitude artifacts remain in regions of rapid phase variation, most visibly at the fifth and seventh echoes. For Stage~2, early-echo phase-evolution fitting of the Stage~1 reconstruction provides an updated phase prior; at the fifth and seventh echoes, the resulting phase estimates more closely resemble the three-average reference (yellow arrows). Stage~3 derives its phase directly from the voxelwise phase of the Stage~2 complex reconstruction. Both later stages show fewer artifacts in rapidly varying phase regions, with fine anatomical detail becoming progressively clearer. Residual noise remains in the three-average reference. Red arrows indicate phase differences between the single- and three-average acquisitions, potentially reflecting time-varying B0 field fluctuations due to system drift and gradient heating during the prolonged scan.

\begin{figure}[t]
    \centering
    \includegraphics[width=0.98\textwidth]{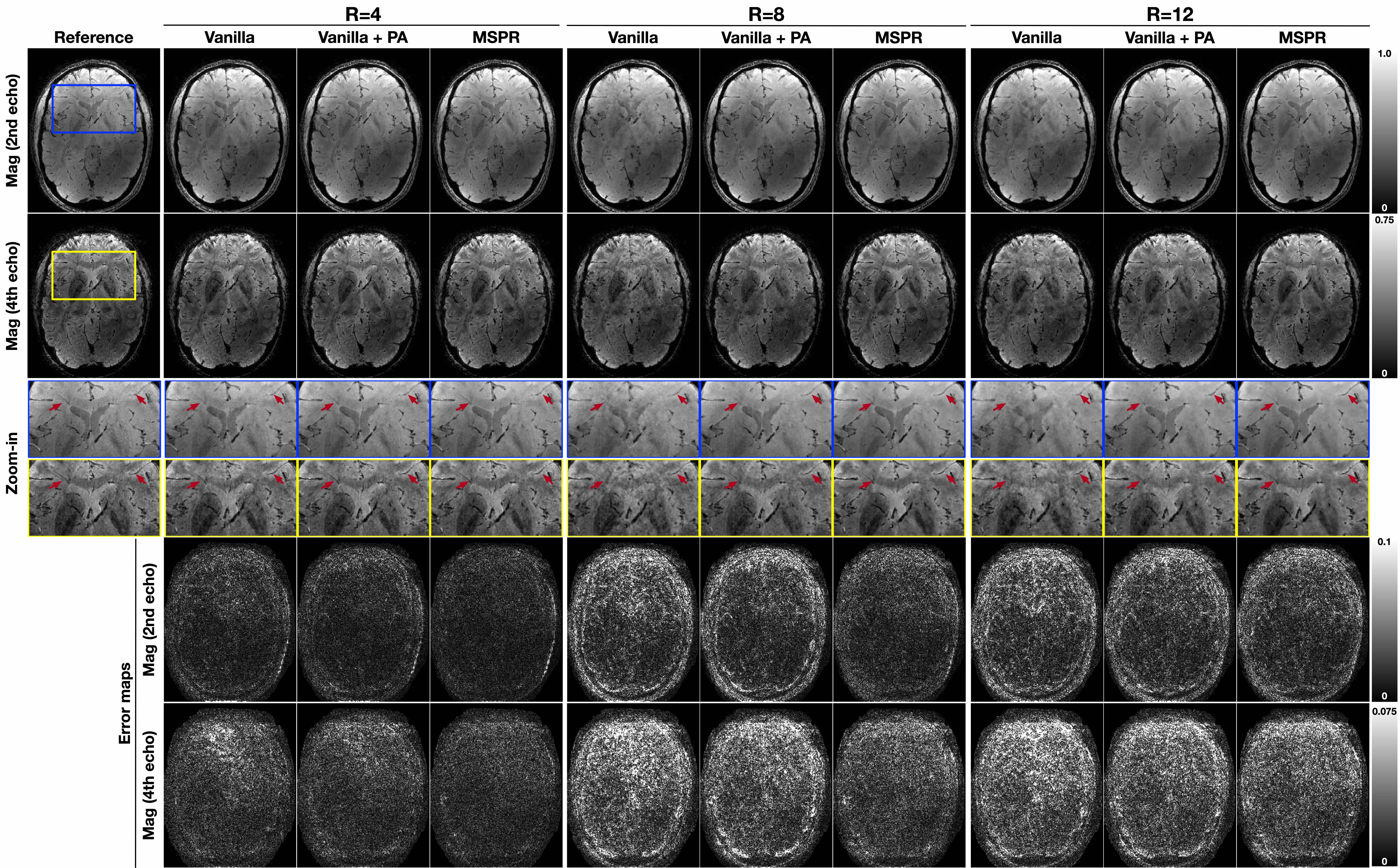}
    \caption{Supervised unrolled-network reconstructions of AHEAD data at acceleration factors $R=4$, $8$, and $12$. Three configurations are compared: Vanilla, a single-stage network without phase-aware reconstruction; Vanilla+PA, a single-stage network with phase-aware reconstruction; and MSPR, a three-stage phase-aware network. Magnitude images from the second and fourth echoes are shown alongside the reference images. Absolute-error maps, representing absolute differences from the reference, are displayed at 10-fold magnification. Blue and yellow boxes mark regions enlarged below, and red arrows highlight fine anatomical structures. Across all evaluated acceleration factors, MSPR achieves the lowest NRMSE and highest PSNR, with the smallest residual errors and improved preservation of anatomical detail.}
    \label{fig:Fig_MGRE_AHEAD_Supervised}
\end{figure}

Figure~\ref{fig:Fig_MGRE_AHEAD_Supervised} compares magnitude reconstructions from three network configurations: conventional single-stage unrolled reconstruction, single-stage phase-aware reconstruction, and three-stage phase-aware reconstruction. The three-stage approach achieved the lowest NRMSE and highest PSNR at acceleration factors of $R=4$, $R=8$, and $R=12$. Conventional reconstruction without phase-aware reconstruction exhibited noticeable blurring and artifacts, particularly at higher acceleration factors. Introducing phase-aware reconstruction in the single-stage network improved anatomical detail and reduced artifacts, while the three-stage approach provided further improvements. Consistently, error maps showed the largest residual errors for conventional reconstruction, followed by single-stage phase-aware reconstruction, with the smallest errors observed for the three-stage approach. Magnified views further highlighted improved preservation of fine anatomical structures.

The corresponding $T_2^*$ and QSM comparisons in Supplementary Figure~\ref{Fig_MGRE_AHEAD_Supervised_t2} demonstrated that three-stage phase-aware reconstruction produced more accurate quantitative maps with fewer visible artifacts and smaller residual errors. These improvements were further supported by the quantitative metrics in Table~\ref{tab:ahead_supervised_metrics}, where the three-stage approach achieved the best overall scores. Together, these results suggest that phase-aware reconstruction improves supervised mGRE reconstruction, with additional benefits from the three-stage framework.

Additional experiments examined the effects of reconstruction stage count and representation (Figures~\ref{fig:Fig_mGRE_numofstages} and~\ref{fig:Fig_mGRE_imagespace_pk}). For the 30-$\mu\mathrm{m}$ dataset, the largest improvement occurred when increasing the model from one to two stages; a third stage yielded smaller additional gains, while a fourth provided little further benefit. Image-domain and three-basis temporal-subspace reconstructions showed visually comparable quality, while the subspace representation reduced the number of coefficient maps from eight to three. Together, these comparisons suggest that additional phase-refinement stages and temporal-subspace modeling offer distinct design trade-offs, although these findings are limited to the evaluated dataset.

\section{Discussion}

MSPR-GRE uses coarse-to-fine reconstruction: Stage~1 uses central support $\Theta^{(1)}$ to form a low-resolution image and initial phase prior, and later stages add higher spatial frequencies while updating phase. This progression is intended to recover fine anatomy from a stable initialization. Results showed improved detail and fewer artifacts in regions of rapid phase variation, although errors remained at later echoes and in complex-phase regions. Overall, progressive k-space expansion with phase-aware reconstruction improved image quality and downstream quantitative mapping.

Explicit phase modeling reduces temporal complexity after demodulation, supporting a relaxation-informed subspace representation. In the self-supervised ablations, phase awareness and early-echo phase fitting reduced visible artifacts; additional stages provided further improvement, although phase accuracy was not directly quantified. In supervised AHEAD experiments, phase-aware reconstruction improved magnitude images and quantitative maps over the non-phase-aware configuration. Together, these results support combining phase modeling with progressive refinement, while future work should evaluate phase accuracy, partial-Fourier acquisitions, pronounced off-resonance, and integration with other network architectures.

The relaxation-informed temporal subspace exploits inter-echo correlations and reduces coefficient maps from eight to three in the eight-echo dataset. Its accuracy depends on the signal model, simulated relaxation and off-resonance ranges, and subspace rank. Susceptibility gradients, intravoxel dephasing, partial-volume effects, and low SNR may cause model mismatch despite flexible complex-valued coefficients. For short echo trains, image-domain reconstruction avoids subspace approximation errors and temporal-basis compression.

The optimal number of stages may depend on phase complexity. In our one-to-four-stage comparison (Supplementary Figure~\ref{fig:Fig_mGRE_numofstages}), most improvement occurred from one to two stages, with smaller gains from a third and little additional benefit from a fourth. Multiple stages may be useful with rapid phase accumulation or substantial $B_0$ inhomogeneity, whereas a single phase-aware stage may suffice for smoother phase. Further evaluation across datasets is needed to balance reconstruction accuracy and computational cost.

This study has limitations. The ex vivo experiments included few specimens, limiting assessment of variability and generalizability, and averaged references may retain noise or phase inconsistencies. The self-supervised ablations were primarily qualitative, while the separate supervised experiments do not establish MSPR-GRE phase accuracy. Subspace performance and stage count may depend on acquisition and phase complexity. In addition, prolonged GRE scans may experience system drift; although inter-acquisition phase differences could reflect time-varying $B_0$ fluctuations, drift was not monitored. Future work should use larger, more diverse datasets; quantify phase accuracy and downstream mapping; compare stage counts and subspace configurations; and investigate navigator-based $B_0$ monitoring and correction.

\section{Conclusions}

We introduced MSPR-GRE, a phase-aware, multi-stage self-supervised framework for accelerated mGRE reconstruction. Progressive phase updates reduced visible artifacts and preserved anatomical detail in ex vivo mouse-brain experiments without fully sampled training targets. In supervised AHEAD experiments, phase-aware multi-stage reconstruction improved image and quantitative-mapping metrics. Future work should assess phase accuracy, generalizability, and performance with other acquisition schemes and architectures.

\subsection*{Funding}
This work was supported in part by the National Institutes of Health under Grant R01 NS125020, the Ralph W. and Grace M. Showalter Research Award, and the UT Rising STARs Award.

\subsection*{Data Availability Statement}
The code supporting this study will be released in a public repository upon publication. Data will be made available from the corresponding authors on reasonable request, subject to applicable ethical and institutional restrictions.

\FloatBarrier

\bibliographystyle{unsrt}
\bibliography{references}

\section*{Supplementary Material}

\setcounter{section}{0}
\renewcommand{\thesection}{S\arabic{section}}
\renewcommand{\theHsection}{S\arabic{section}}

\setcounter{figure}{0}
\setcounter{table}{0}
\renewcommand{\thefigure}{S\arabic{figure}}
\renewcommand{\thetable}{S\arabic{table}}

\section{Temporal Basis Construction}
\label{app:svd_basis}

After demodulation of the dominant phase evolution, the residual GRE signal is approximated by an exponential $T_2^*$ decay with a small residual off-resonance frequency. For echo time $t_n$, relaxation rate $R_j$, and residual off-resonance frequency $\delta f_q$, the simulated signal is defined as

\begin{equation}
    d_n(R_j,\delta f_q)
    =\exp\left[(-R_j+i2\pi\delta f_q)t_n\right].
    \label{eq:app_signal_dictionary}
\end{equation}

A signal dictionary $\mathbf M\in\mathbb C^{N_e\times JQ}$ is constructed from simulated signal evolutions spanning prescribed relaxation rates $\{R_j\}_{j=1}^J$ and residual off-resonance frequencies $\{\delta f_q\}_{q=1}^Q$. Singular value decomposition (SVD) is then applied to extract the leading temporal basis functions \cite{tamir2017t2shuffling,dong2020epti}:

\begin{equation}
    \mathbf M=\mathbf U\boldsymbol{\Sigma}\mathbf V^H,
    \qquad
    \boldsymbol{\phi}=\mathbf U_{:,1:L},
    \label{eq:app_basis}
\end{equation}

where $\boldsymbol{\phi}\in\mathbb C^{N_e\times L}$ contains the first $L$ left singular vectors and satisfies $\boldsymbol{\phi}^H\boldsymbol{\phi}=\mathbf I_L$.

For MSPR-GRE, the relaxation-informed temporal basis is constructed from simulated GRE signal evolutions spanning $T_2^*$ values of $5$--$150$~ms and residual off-resonance frequencies of $-10$ to $10$~Hz. The $T_2^*$ range was selected to broadly encompass relaxation characteristics observed in high-field human brain and ex vivo mouse-brain imaging, while allowing for rapidly decaying signals and variability across tissue types~\cite{peters2007t2,caan2019mp2rageme}. Since the dominant off-resonance-induced phase evolution is explicitly modeled by the phase prior, the temporal basis only needs to represent the remaining phase evolution. Accordingly, a narrower residual off-resonance range of $\pm10$~Hz was adopted, consistent with the use of reduced residual frequency ranges following field-map refinement in subspace-based reconstruction~\cite{wang2025spherical}. The three leading temporal basis functions are retained for subspace reconstruction.

As defined in Eq.~\eqref{eq:stage_basis}, Stage~1 and acquisitions with $N_e\leq3$ use the identity basis $\boldsymbol{\phi}_s=\mathbf I_{N_e}$, corresponding to image-domain reconstruction without temporal-subspace compression. For $N_e\geq4$, Stages~2 and~3 use the SVD-derived basis $\boldsymbol{\phi}$. The phase evolution is modeled separately by $\mathbf B_{s-1}$, as described in the main text.

\section{Phase-Informed LLR-Regularized Subspace Reconstruction}
We formulate the phase-informed, LLR-regularized subspace reconstruction as the following optimization problem:

\begin{equation}
\begin{aligned}
    \widehat{\mathbf{c}}
    =\underset{\mathbf{c}}{\operatorname{arg\,min}}\quad&
    \left\|
    \mathbf{A}\mathbf{B}
    \boldsymbol{\phi}\mathbf{c}
    -\mathbf{y}
    \right\|_2^2\\
    &+\mu\mathcal{L}(R_r\mathbf{c}),
\end{aligned}
\label{eq:llr_recon}
\end{equation}

Here, $\mathbf{A}$ denotes the multi-coil encoding operator, $\mathbf{B}$ applies the phase prior, $\boldsymbol{\phi}$ is the temporal-subspace basis, and $\mathbf{c}$ contains the spatial coefficient maps. For each local neighborhood indexed by $r$, $R_r$ extracts the corresponding coefficient patches and arranges them as a Casorati matrix. The regularizer $\mathcal{L}$ promotes low rank in these local matrices, with $\mu$ balancing data fidelity and regularization.

We implemented in MATLAB a two-stage phase-informed subspace reconstruction. Both stages use the full acquired k-space support. We estimated the initial low-resolution phase prior from the central $0.5\times0.5\times0.5$ fraction of k-space along the three spatial dimensions. After apodization with a smooth window, these data were transformed using a 3D inverse Fourier transform and combined across coils using sensitivity weighting to obtain low-resolution multi-echo images. We extracted the phase of these images to form the initial complex-valued phase prior.

We derived the temporal basis by applying SVD to a dictionary of simulated complex-valued multi-echo GRE signals. The forward model combines temporal-basis expansion and the spatially and echo-dependent phase prior with coil-sensitivity and 3D Fourier encoding, restricted to the acquired sampling mask.

We estimated the subspace coefficient images by minimizing a k-space data-fidelity objective with LLR regularization. LLR regularization was applied to the complex-valued coefficients in nonoverlapping $6\times6\times6$ spatial patches using singular-value soft-thresholding. We optimized the objective with the alternating direction method of multipliers (ADMM), using conjugate-gradient (CG) iterations for the data-consistency update.

After Stage~1, we refined the phase prior using the intermediate complex multi-echo reconstruction. We removed the initial phase contribution and estimated the residual echo-dependent phase relative to a reference echo. Spatial smoothing and signal-reliability weighting were used to limit phase-estimation errors in low-SNR regions. We combined the estimated residual phase with the initial prior to form the updated phase estimate for Stage~2. The LLR regularization weight, $\mu$, was tuned empirically to balance noise suppression against preservation of fine anatomical detail.

\section{Unrolled Network Architecture}

\begin{figure}[htbp]
    \centering
    \includegraphics[width=0.9\textwidth]{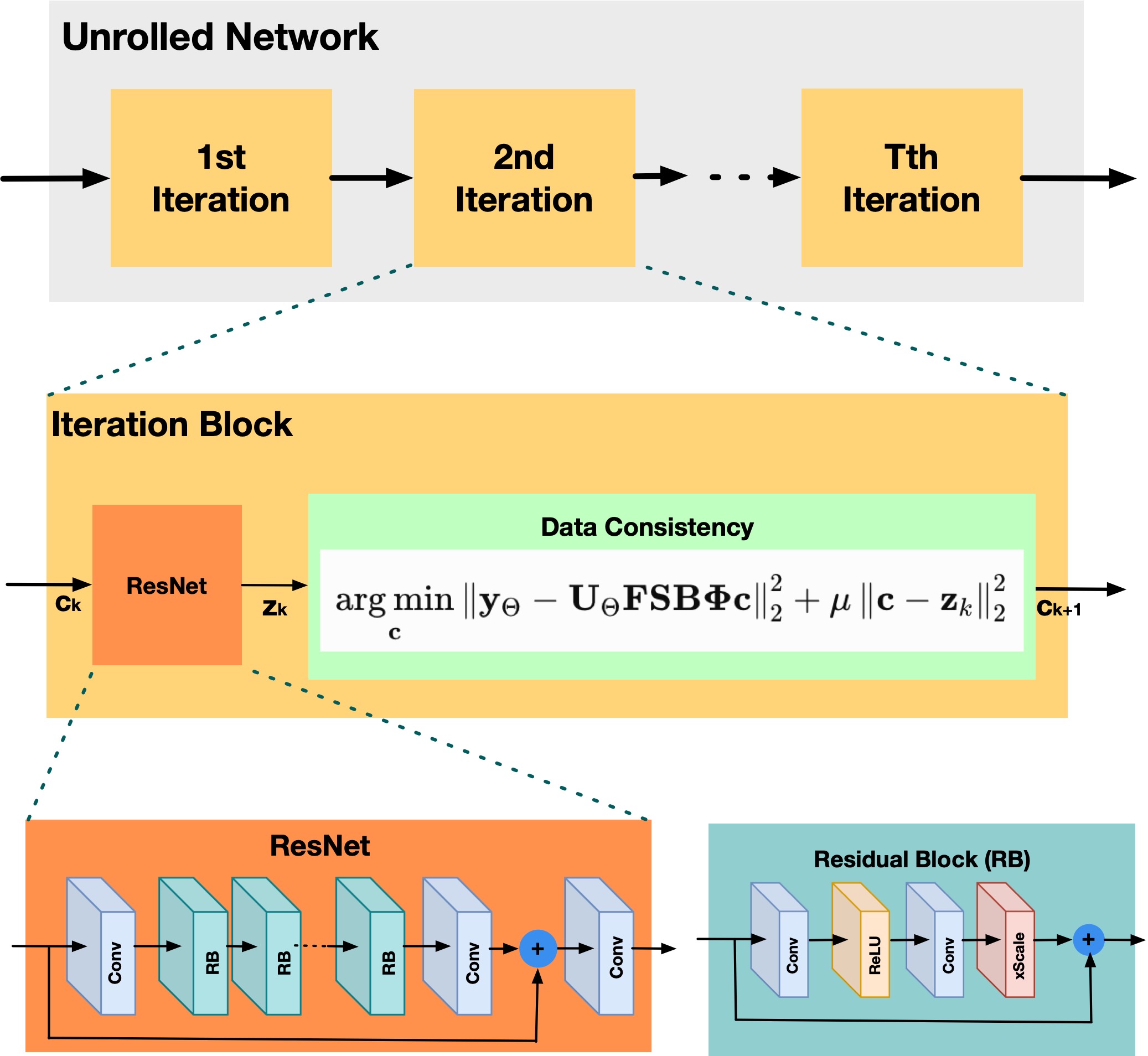}
    \caption{Schematic of the unrolled reconstruction network and its iteration block. Each iteration maps the current estimate $c_k$ through a ResNet to obtain $z_k$, then applies a data-consistency update that balances agreement with acquired k-space samples against proximity to $z_k$, producing $c_{k+1}$. The ResNet comprises convolutional layers and residual blocks; each residual block applies convolution, ReLU, convolution, and a learned scaling factor before adding a skip connection. In the displayed objective, $y_\Theta$ denotes acquired k-space data, $\mathcal{U}_\Theta$ the sampling operator, $F$ the Fourier transform, $S$ the coil-sensitivity operator, $B$ the phase operator, $\phi$ the temporal-subspace basis, and $\mu$ the regularization weight.}
    \label{fig:Fig_mGRE_unrollednetwork}
\end{figure}

The optimization problem in Eq.~\eqref{eq:unified_recon} is solved using variable splitting with a quadratic-penalty formulation, following the approach used in SSDU \cite{yaman2020ssdu}. The resulting iterations are unrolled into a network comprising alternating learned-regularization and data-consistency blocks (Figure~\ref{fig:Fig_mGRE_unrollednetwork}).

Each learned-regularization block uses a ResNet, and data consistency is enforced using conjugate gradients (CG). The real and imaginary components of the multi-echo images or temporal-subspace coefficient maps are concatenated along the channel dimension and processed jointly by the ResNet. Across all experiments, each ResNet contains five residual blocks, and each data-consistency update uses five CG iterations.

\section{Additional Results for the 18-$\mu$m Dataset}
\label{app:18um_more_result}

Table~\ref{tab:dataset2_metrics} presents the quantitative evaluation of single-average reconstructions at 18~$\mu$m isotropic resolution, using the four-average reconstruction as the reference. However, the four-average reference exhibits substantial residual noise, potentially arising from low-SNR, system drift, gradient heating, and intra-scan variations during the prolonged acquisition. These imperfections compromise its reliability as a reference and may introduce biases into reference-based comparisons. Consequently, the reported quantitative metrics should be interpreted with caution, as they may not fully reflect the actual reconstruction quality or the preservation of fine anatomical structures.



\begin{table}[t]
    \centering
    \caption{Quantitative reconstruction metrics for the 18-$\mu$m dataset using the four-average reconstruction as the reference. Bold indicates the best score for each output, acceleration, and metric (lowest NRMSE; highest PSNR and SSIM).}
    \label{tab:dataset2_metrics}
    \small
    \setlength{\tabcolsep}{4pt}
    \renewcommand{\arraystretch}{1.05}
    \begin{tabular*}{\textwidth}{@{\extracolsep{\fill}}llrrrrrr@{}}
        \toprule
        \textbf{Output} & \textbf{Method} & \multicolumn{3}{c}{$R=1$} & \multicolumn{3}{c}{$R=3$} \\
        \cmidrule(lr){3-5}\cmidrule(lr){6-8}
        & & \textbf{NRMSE (\%)$\downarrow$} & \textbf{PSNR $\uparrow$} & \textbf{SSIM $\uparrow$} & \textbf{NRMSE (\%)$\downarrow$} & \textbf{PSNR $\uparrow$} & \textbf{SSIM $\uparrow$} \\
        \midrule
        Magnitude, 1st echo 
         & Raw & 16.11 & 29.08 & 0.966 & --- & --- & --- \\
         & Zero-Filled & --- & --- & --- & 15.14 & 29.62 & 0.948 \\
         & TV & \textbf{12.27} & \textbf{31.45} & \textbf{0.975} & \textbf{12.19} & \textbf{31.50} & \textbf{0.972} \\ 
         & SSDU & 13.52 & 30.60 & 0.969 & 14.26 & 30.14 & 0.964 \\
         & MSPR-GRE & 13.02 & 30.93 & 0.973 & 13.23 & 30.79 & 0.970 \\
        \addlinespace
        Magnitude, 2nd echo 
         & Raw & 31.87 & 24.88 & 0.946 & --- & --- & --- \\
         & Zero-Filled & --- & --- & --- & 26.90 & 26.35 & 0.936 \\
         & TV & 22.55 & 27.89 & \textbf{0.960} & \textbf{22.38} & \textbf{27.95} & \textbf{0.953} \\  
         & SSDU & 23.60 & 27.49 & 0.955 & 24.46 & 27.18 & 0.947 \\
         & MSPR-GRE & \textbf{22.42} & \textbf{27.93} & 0.957 & 22.88 & 27.76 & 0.953 \\
        \addlinespace
        QSM  
         & Raw & \textbf{81.71} & \textbf{26.49} & \textbf{0.940} & --- & --- & --- \\
         & Zero-Filled & --- & --- & --- & 87.46 & 25.90 & 0.842 \\
         & TV & 84.89 & 26.16 & 0.852 & 86.94 & 25.95 & 0.843 \\   
         & SSDU & 83.96 & 26.25 & 0.858 & \textbf{85.50} & \textbf{26.09} & \textbf{0.852} \\
         & MSPR-GRE & 83.62 & 26.29 & 0.861 & 85.94 & 26.05 & \textbf{0.852} \\
        \bottomrule
    \end{tabular*}
\end{table}

\section{Effects of Phase-Aware Reconstruction on Supervised Learning}
\label{app:supervised_training}

\begin{figure}[t]
    \centering
    \includegraphics[width=\textwidth]{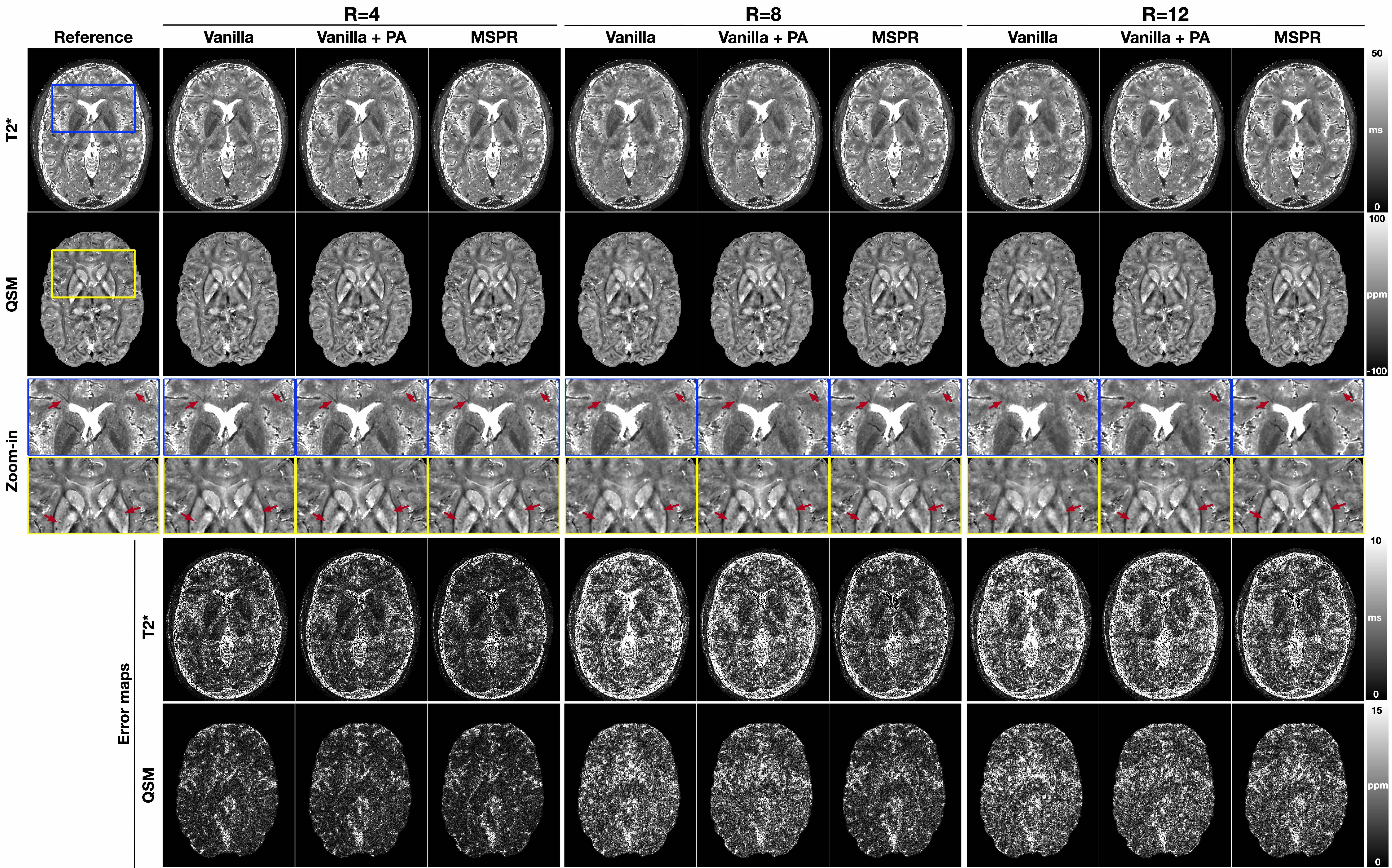}
    \caption{Comparison of supervised unrolled-network reconstructions of AHEAD data with and without phase-aware reconstruction, including $T_2^*$ and QSM maps. Three network configurations are evaluated: Vanilla, a single-stage unrolled network without phase-aware reconstruction; Vanilla+PA, a single-stage unrolled network with phase-aware reconstruction; and MSPR, a three-stage phase-aware reconstruction model. Reconstructed quantitative maps are compared with the reference reconstruction at acceleration factors of $R=4$, $R=8$, and $R=12$. Difference maps show absolute errors relative to the reference and are displayed on a 10-fold intensity scale. Blue and yellow boxes indicate regions enlarged in the bottom rows, while red arrows highlight fine anatomical structures. MSPR demonstrates improved reconstruction quality and more accurate quantitative maps, with the smallest differences relative to the reference among the evaluated methods.}
    \label{Fig_MGRE_AHEAD_Supervised_t2}
\end{figure}

To evaluate the benefits of phase-aware reconstruction in a supervised learning setting, we compared three unrolled network configurations using the AHEAD dataset. All models were evaluated on retrospectively undersampled data at acceleration factors of $R=4$, $8$, and $12$. Figure~\ref{Fig_MGRE_AHEAD_Supervised_t2} presents the reconstructed $T_2^*$ and QSM maps, while the corresponding magnitude images are shown in Figure~\ref{fig:Fig_MGRE_AHEAD_Supervised}. Across the evaluated acceleration factors, the multi-stage phase-aware reconstruction (MSPR) model generally produced quantitative maps that were more accurate and had lower reconstruction errors relative to the reference than those from the single-stage configurations. These findings suggest that incorporating phase information into the reconstruction model, together with a multi-stage refinement strategy, can improve downstream quantitative mapping in supervised mGRE reconstruction. Table~\ref{tab:ahead_supervised_metrics} summarizes the quantitative results, showing that MSPR achieved the best overall performance across the evaluated acceleration factors.

\begin{table}[!htbp]
    \centering
    \caption{Quantitative scores for reconstructions of the AHEAD dataset (mean $\pm$ standard deviation). Bold indicates the best mean score for each output, acceleration, and metric (lowest NRMSE; highest PSNR and SSIM).}
    \label{tab:ahead_supervised_metrics}
    \small
    \setlength{\tabcolsep}{4pt}
    \renewcommand{\arraystretch}{1.05}
    \resizebox{\textwidth}{!}{%
    \begin{tabular}{@{}llrrrrrrrrr@{}}
        \toprule
        \textbf{Output} & \textbf{Method} & \multicolumn{3}{c}{$R=4$} & \multicolumn{3}{c}{$R=8$} & \multicolumn{3}{c}{$R=12$} \\
        \cmidrule(lr){3-5}\cmidrule(lr){6-8}\cmidrule(lr){9-11}
        & & \textbf{NRMSE (\%) $\downarrow$} & \textbf{PSNR $\uparrow$} & \textbf{SSIM $\uparrow$} & \textbf{NRMSE (\%) $\downarrow$} & \textbf{PSNR $\uparrow$} & \textbf{SSIM $\uparrow$} & \textbf{NRMSE (\%) $\downarrow$} & \textbf{PSNR $\uparrow$} & \textbf{SSIM $\uparrow$}\\
        \midrule
        Magnitude, 2nd echo
         & Vanilla & 
         $5.09 \pm 0.22$ & $38.62 \pm 0.86$ & $0.9517 \pm 0.0071$ & 
         $8.94 \pm 0.40$ & $33.72 \pm 0.86$ & $0.8870 \pm 0.0143$ & 
         $8.90 \pm 0.43$ & $33.77 \pm 0.85$ & $0.8829 \pm 0.0158$ \\

         & Vanilla + PA & 
         $4.95 \pm 0.27$ & $38.86 \pm 1.04$ & $0.9553 \pm 0.0075$ & 
         $8.44 \pm 0.50$ & $34.24 \pm 1.05$ & $0.9014 \pm 0.0147$ & 
         $8.36 \pm 0.46$ & $34.31 \pm 0.98$ & $0.8949 \pm 0.0161$ \\

         & MSPR & 
         $\mathbf{4.05} \pm 0.30$ & $\mathbf{40.61} \pm 1.09$ & $\mathbf{0.9655} \pm 0.0058$ & 
         $\mathbf{5.55} \pm 0.39$ & $\mathbf{37.88} \pm 1.08$ & $\mathbf{0.9402} \pm 0.0107$ & 
         $\mathbf{7.06} \pm 0.41$ & $\mathbf{35.78} \pm 1.03$ & $\mathbf{0.9159} \pm 0.0143$ \\
        \addlinespace

        Magnitude, 4th echo
         & Vanilla & 
         $8.30 \pm 0.48$ & $36.93 \pm 0.96$ & $0.9417 \pm 0.0082$ & 
         $16.09 \pm 0.90$ & $31.18 \pm 0.91$ & $0.8442 \pm 0.0163$ & 
         $15.25 \pm 0.71$ & $31.65 \pm 0.87$ & $0.8421 \pm 0.0171$ \\

         & Vanilla + PA & 
         $8.00 \pm 0.46$ & $37.26 \pm 0.94$ & $0.9461 \pm 0.0082$ & 
         $15.32 \pm 0.88$ & $31.61 \pm 0.92$ & $0.8587 \pm 0.0165$ & 
         $14.30 \pm 0.77$ & $32.21 \pm 0.92$ & $0.8576 \pm 0.0181$ \\

         & MSPR & 
         $\mathbf{6.57} \pm 0.39$ & $\mathbf{38.97} \pm 0.88$ & $\mathbf{0.9582} \pm 0.0068$ 
         & $\mathbf{9.24} \pm 0.58$ & $\mathbf{36.00} \pm 0.89$ & $\mathbf{0.9207} \pm 0.0123$ 
         & $\mathbf{12.19} \pm 0.56$ & $\mathbf{33.59} \pm 0.77$ & $\mathbf{0.8811} \pm 0.0153$ \\
        \addlinespace

        $T_2^*$
         & Vanilla & 
         $41.05 \pm 2.31$ & $22.51 \pm 0.91$ & $0.8934 \pm 0.0175$ & 
         $64.84 \pm 1.31$ & $18.53 \pm 0.85$ & $0.7557 \pm 0.0298$ & 
         $66.60 \pm 1.66$ & $18.98 \pm 0.86$ & $0.7679 \pm 0.0296$ \\

         & Vanilla + PA & 
         $38.75 \pm 2.70$ & $23.02 \pm 0.90$ & $0.9055 \pm 0.0160$ & 
         $58.78 \pm 2.66$ & $19.39 \pm 0.78$ & $0.7874 \pm 0.0256$ & 
         $55.89 \pm 2.64$ & $19.83 \pm 0.86$ & $0.8006 \pm 0.0266$ \\

         & MSPR & 
         $\mathbf{32.47} \pm 2.78$ & $\mathbf{24.56} \pm 1.03$ & $\mathbf{0.9343} \pm 0.0124$ & 
         $\mathbf{42.16} \pm 2.52$ & $\mathbf{22.28} \pm 0.92$ & $\mathbf{0.8850} \pm 0.0186$ & 
         $\mathbf{50.62} \pm 2.24$ & $\mathbf{20.69} \pm 0.85$ & $\mathbf{0.8368} \pm 0.0213$ \\
        \addlinespace

        QSM
         & Vanilla & 
         $37.34 \pm 1.49$ & $38.64 \pm 0.90$ & $0.9321 \pm 0.0099$ & 
         $52.13 \pm 2.29$ & $35.75 \pm 0.92$ & $0.8320 \pm 0.0243$ & 
         $53.01 \pm 2.267$ & $35.60 \pm 0.87$ & $0.8258 \pm 0.0243$ \\

         & Vanilla + PA & 
         $36.90 \pm 1.30$ & $38.74 \pm 0.86$ & $0.9358 \pm 0.0089$ & 
         $50.61 \pm 2.06$ & $36.00 \pm 0.88$ & $0.8483 \pm 0.0216$ & 
         $51.18 \pm 1.87$ & $35.90 \pm 0.81$ & $0.8459 \pm 0.0217$ \\

         & MSPR & 
         $\mathbf{34.82} \pm 1.08$ & $\mathbf{39.25} \pm 0.81$ & $\mathbf{0.9471} \pm 0.0059$ & 
         $\mathbf{42.03} \pm 1.59$ & $\mathbf{37.61} \pm 0.85$ & $\mathbf{0.9128} \pm 0.0117$ &
         $\mathbf{50.99} \pm 2.03$ & $\mathbf{35.94} \pm 0.86$ & $\mathbf{0.8550} \pm 0.0185$ \\
        \bottomrule
    \end{tabular}%
    }
\end{table}

\section{Effect of the Number of Reconstruction Stages}

We compared MSPR-GRE models with one, two, three, or four reconstruction stages on the eight-echo, 30-$\mu\mathrm{m}$ dataset. Visually, the single-stage model retained artifacts in regions of strong phase variation, whereas the multi-stage reconstructions reduced these artifacts and appeared similar to one another. Quantitative metrics showed a substantial improvement from one to two stages: NRMSE decreased from 17.06\% to 14.79\% for the third echo and from 24.29\% to 21.16\% for the fifth echo. Adding a third stage further reduced NRMSE to 14.27\% and 20.46\%, respectively, and increased PSNR to 31.34 and 29.15 dB. A fourth stage yielded only small, mixed changes: third-echo NRMSE decreased to 14.12\%, while fifth-echo NRMSE increased slightly to 20.52\%. SSIM changed little beyond two stages. Thus, most of the observed improvement occurred by two stages, with a further modest gain at three stages and no consistent additional benefit from the fourth stage in this dataset.

Because phase complexity varies across acquisitions, the optimal number of stages may also be data dependent. A single stage may suffice when phase is relatively smooth, whereas more complex phase patterns may benefit from additional refinement. This possibility requires systematic evaluation across datasets with differing phase characteristics.

\begin{figure}[!htbp]
    \centering
    \includegraphics[width=\textwidth]{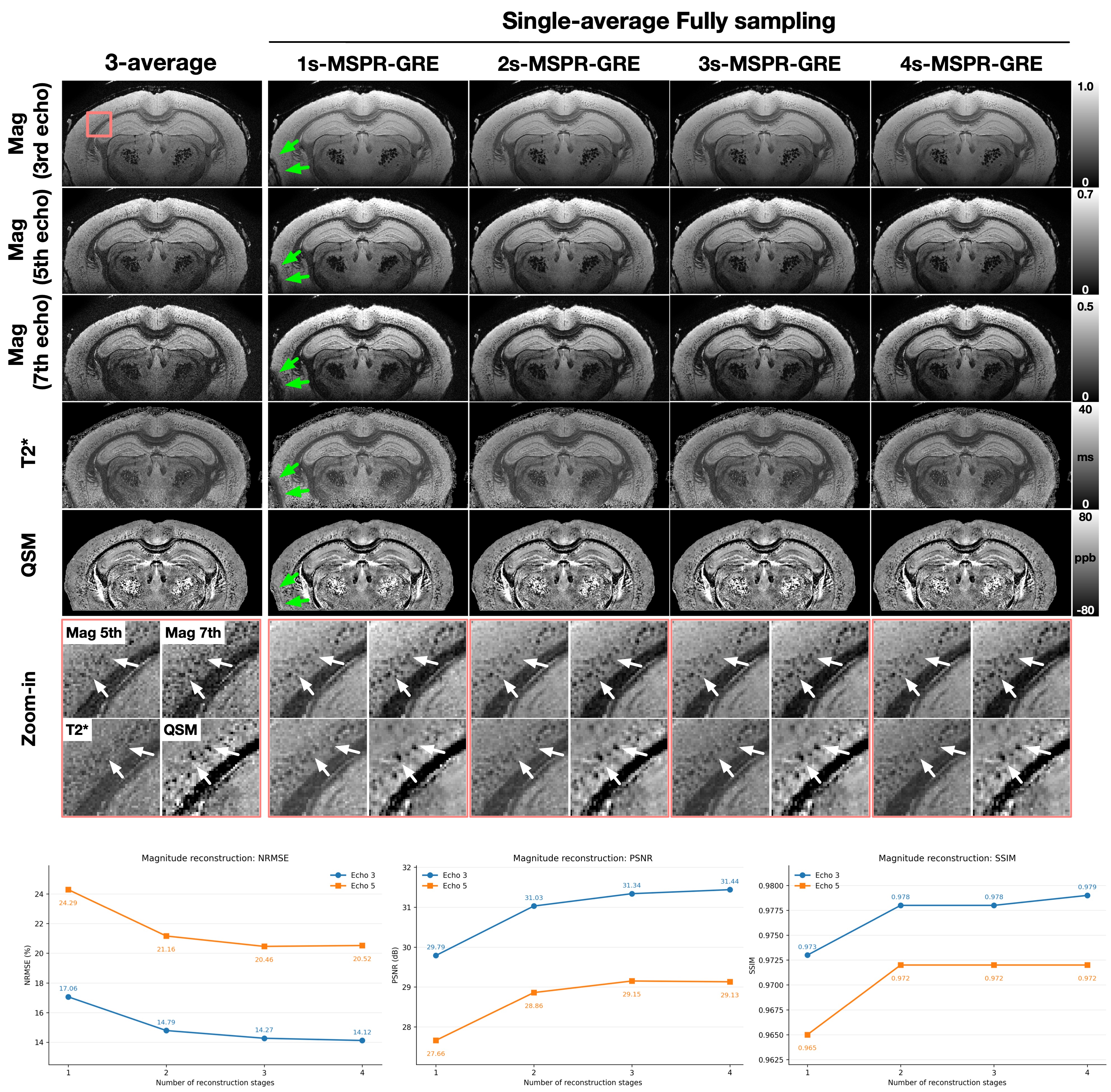}
    \caption{Visual and quantitative comparison of MSPR-GRE models with one, two, three, and four stages for the eight-echo, 30-$\mu\mathrm{m}$ ex vivo mouse-brain dataset. The left column shows the three-average reference; the remaining columns show the reconstructions from each model. Green arrows mark regions of strong phase variation, where artifacts are visible in the single-stage result and reduced in the multi-stage results. The multi-stage reconstructions appear visually similar. Quantitative NRMSE, PSNR, and SSIM for the third- and fifth-echo magnitude images are also shown.}
    \label{fig:Fig_mGRE_numofstages}
\end{figure}

\begin{figure}[!htbp]
    \centering
    \includegraphics[width=\textwidth]{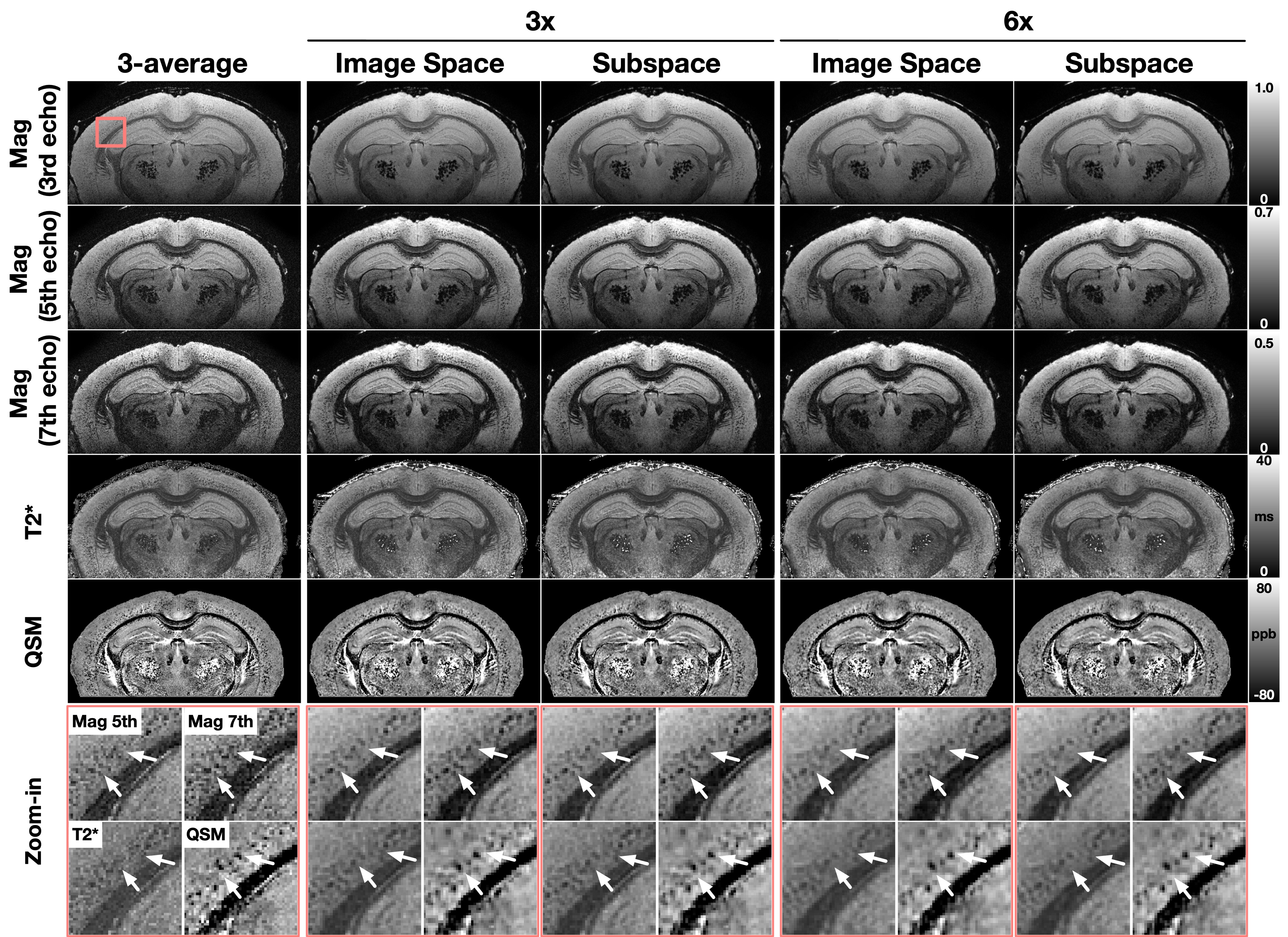}
    \caption{Comparison of image-domain and three-basis temporal-subspace MSPR-GRE reconstructions for the eight-echo, 30-$\mu\mathrm{m}$ ex vivo mouse-brain dataset. The left column shows the three-average reference; the remaining columns show reconstructions from retrospectively undersampled data at acceleration factors of $R=3$ and $R=6$. Rows show magnitude images at the third, fifth, and seventh echoes, followed by $T_2^*$ and QSM maps. Bottom panels magnify the red-boxed region, with white arrows indicating fine anatomical structures. The two representations yield visually comparable results, while the subspace model reduces the number of coefficient maps from eight to three (62.5\%).}
    \label{fig:Fig_mGRE_imagespace_pk}
\end{figure}

\section{Temporal-Subspace Versus Image-Domain Reconstruction}
We compared MSPR-GRE with direct image-domain reconstruction at all stages against a variant using a three-basis temporal subspace in Stages~2 and~3 for the eight-echo, 30-$\mu\mathrm{m}$ dataset. Visual inspection showed comparable magnitude-image quality and preservation of fine anatomical structures across echoes. At $R=3$, the subspace variant slightly improved mean magnitude NRMSE across the eight echoes (19.50\% to 19.27\%), PSNR (30.17 to 30.27~dB), and SSIM (0.9714 to 0.9723). At $R=6$, mean magnitude NRMSE was nearly unchanged (19.81\% for image-domain reconstruction and 19.84\% for subspace reconstruction). QSM NRMSE was also similar: 68.10\% for image-domain versus 68.06\% for subspace reconstruction at $R=3$, and 71.15\% versus 70.93\% at $R=6$. The temporal-subspace representation reduces the number of coefficient maps from eight to three (62.5\%) in the subspace stages. Overall, these results indicate that the three-basis representation provides a more compact model while maintaining reconstruction metrics and visual quality comparable to direct image-domain reconstruction in this dataset.



\end{document}